\documentclass[number,preprint,3p]{elsarticle}

\usepackage{color}
\usepackage{graphicx}
\usepackage{subcaption}
\usepackage{algorithm}
\usepackage{bm}

\pdfoutput=1

\usepackage{hyperref}

\hypersetup{
    colorlinks=true,
    linkcolor=black,
    citecolor=black,
    filecolor=black,
    urlcolor=black,
}

\usepackage{amssymb}
\usepackage{amsthm}
\usepackage{amsmath}
\usepackage{booktabs}
\usepackage{url}
\usepackage{scrextend}
\usepackage{epstopdf}
\usepackage{float}
\usepackage[perpage]{footmisc}

\makeatletter
\newenvironment{breakablealgorithm}
  {
   \begin{center}
     \refstepcounter{algorithm}
     \hrule height.8pt depth0pt \kern2pt
     \renewcommand{\caption}[2][\relax]{
       {\raggedright\textbf{\ALG@name~\thealgorithm} ##2\par}%
       \ifx\relax##1\relax 
         \addcontentsline{loa}{algorithm}{\protect\numberline{\thealgorithm}##2}%
       \else 
         \addcontentsline{loa}{algorithm}{\protect\numberline{\thealgorithm}##1}%
       \fi
       \kern2pt\hrule\kern2pt
     }
  }{
     \kern2pt\hrule\relax
   \end{center}
  }
\makeatother

\biboptions{square}
\journal{arXiv}
\begin{document}

\begin{frontmatter}



\title{A novel active learning-based Gaussian process metamodelling strategy for estimating the full probability distribution in forward UQ analysis}

 \author[1]{Ziqi Wang}
 \author[2]{Marco Broccardo}
 \address[1]{Earthquake Engineering Research and Test Center, Guangzhou University, China}
 \address[2]{Swiss Seismological Service, SED, ETH Z\"urich, Switzerland}

\begin{abstract}
This paper proposes an active learning-based Gaussian process (AL-GP)  metamodelling method to estimate the cumulative as well as complementary cumulative distribution function (CDF/CCDF) for forward uncertainty quantification (UQ) problems. Within the field of UQ, previous studies focused on developing AL-GP approaches for reliability (rare event probability) analysis of expensive black-box solvers. A naive iteration of these algorithms with respect to different CDF/CCDF threshold values would yield a discretized CDF/CCDF. However, this approach inevitably leads to a trade off between accuracy and computational efficiency since both depend (in opposite way) on the selected discretization.
In this study, a specialized error measure and a learning function are developed such that the resulting AL-GP method is able to efficiently estimate the CDF/CCDF for a specified range of interest without an explicit dependency on discretization. Particularly, the proposed AL-GP method is able to simultaneously provide accurate CDF and CCDF estimation in their median-low probability regions. Three numerical examples are introduced to test and verify the proposed method.
\end{abstract}


\end{frontmatter}

\renewcommand{\thefootnote}{\fnsymbol{footnote}}

\section{Introduction}

\noindent
 In a broad sense, uncertainty quantification (UQ) refers to the theory and practice to obtain quantitative understanding on the influences of uncertainties present within computational or real physical models. An incomplete list of possible UQ analysis includes the prediction of probability \cite{Ditlevsen1996}, the interpolation/extrapolation for the most likely outcome \cite{BeckSysId}, the validation/calibration/correction of computational model \cite{LUCAS20084591}, etc. There are intrinsic connections between various branches of UQ, and attempts are made to develop unified UQ frameworks \cite{ReliabilitySIAM}\cite{OUQ}. This study focuses on a central problem in forward UQ problems, the estimation of distribution function, i.e. cumulative and complementary cumulative distribution function (CDF/CCDF). Specifically, consider a system with a finite set of basic random variables $\bm{X}=[X_1,X_2,...,X_n]$ representing the source of randomness. Given the joint probability distribution of $\bm{X}$, we are interested in the probability distribution of a system output, $Y$, propagated from $\bm{X}$. The output $Y$ can be any variables selected to describe the system performance or behavior of interest. The deterministic mapping from $\bm{X}$ to $Y$ is written as
\begin{equation}
Y=\mathcal{M}\left(\bm{X}\right)\, .\label{ModelF}
\end{equation}
\noindent In a more general setting $Y$ can also be a vector of random variables, yet this study focuses (without loss of generality) on the case where $Y$ is unidimensional. Although Eq.\eqref{ModelF} looks trivial, the model function $\mathcal{M}(\cdot)$ may involve computationally expensive models, e.g. finite element algorithms.

Using Eq.\eqref{ModelF}, the CDF of $Y$, denoted by $F_Y(y)$, can be expressed as

\begin{equation}
F_Y(y)\equiv \mathbb{P}\left(Y\leq y\right)=\mathbb{P}\left(\mathcal{M}\left(\bm{X}\right)\leq y\right)=\int_{\mathcal{M}\left(\bm{x}\right)\leq y}f_{\bm{X}}(\bm{x})\,d\bm{x}\,,\label{CDF}
\end{equation}

\noindent where $\mathbb{P}(\cdot)$ denotes probability, $f_{\bm{X}}(\bm{x})$ denotes the joint probability density function (PDF) of the basic random variables $\bm{X}$. The CCDF of $Y$, denoted by $\bar F_Y(y)$ can be expressed as

\begin{equation}
\bar F_Y(y)\equiv \mathbb{P}\left(Y> y\right)=\int_{\mathcal{M}\left(\bm{x}\right)>y}f_{\bm{X}}(\bm{x})\,d\bm{x}=1-F_Y(y)\,.\label{CCDF}
\end{equation}
In terms of the trivial connection between Eq.\eqref{CDF} and Eq.\eqref{CCDF}, it seems given the CDF $F_Y(y)$ one could compute the CCDF simply by $\bar F_Y(y)=1-F_Y(y)$. However, to obtain satisfactory accuracy in the low probability region of the CCDF, instead of using $1-F_Y(y)$, one typically needs to solve the integration in Eq.\eqref{CCDF}\footnote{Observe that we implicitly assume non-symmetric distributions or distributions for which are \textit{not} known a-priory whether they are symmetric or not}. Moreover, note that by definition the low probability region of the CDF is in the left tail of $f_Y(y)$, while the low probability region of the CCDF is in the right tail. Therefore, one typically needs to separately run numerical integration algorithms to capture the left and right tails so that the low probability regions of the CDF and CCDF can be accurately estimated. For non-trivial problems ( e.g. $\mathcal{M}(\cdot)$ involves computationally expensive models or/and the dimensionality of $\bm{X}$ is large) analytical solution of Eq.\eqref{CDF} or Eq.\eqref{CCDF} is typically unfeasible, and deterministic cubature \cite{NumInt}, metamodelling \cite{PCE} and Monte Carlo simulation techniques \cite{MCS} are commonly used to estimate the integration.

In the field of reliability analysis, an active learning-based Gaussian process (AL-GP) metamodelling strategy \citep{echard2011ak} was recently introduced  to estimate rare event probabilities. The approach has proven a remarkable success in solving low-medium dimensional reliability problems. In the context of Eq.\eqref{CDF}, the AL-GP approach for reliability analysis can be interpreted as to train a Gaussian process model to perform interpolation/extrapolation within the domain defined by $G(\bm x|y)=\mathcal{M}\left(\bm{x}\right)-y\leq 0$, so that the probability $\mathbb P(G(\bm x|y)\leq 0)$ can be estimated by the metamodel. Clearly, $\mathbb P(G(\bm x|y)\leq 0)$ as a function of $y$ is by definition the CDF of $Y$. Therefore, the AL-GP approach can be directly used to estimate the CDF for a specified threshold $y$. Similarly, by manipulating $G(\bm x|y)$ the AL-GP approach can also be used to estimate the CCDF. However, if one is interested in the CDF/CCDF corresponding to a relatively wide range of $y$, a naive iteration of the conventional AL-GP method for a sequence of $y$ values (i.e. a discrete mesh)  would lead to a trade-off between accuracy and computational efficiency. Within this research gap, this study introduces a novel ``mesh free'' global AL-GP strategy to estimate the CDF/CCDF. The proposed AL-GP strategy possesses an attractive property such that the median-low probability regions of both CDF and CCDF can be simultaneously estimated.

The paper is organized as follows. Section \ref{Framework} provides a basic framework of the global AL-GP method to estimate the distribution function. Section \ref{Details} provides theoretical and technical details for the proposed method.  Section \ref{Numerical}: first, it introduces a toy example with semi-analytical CDF/CCDF solution to gain a deep understanding of the proposed method; second, it shows an example with highly nonlinear analytical model function; and third, it presents a structural dynamics example with hysteretic force-deformation behavior. Section \ref{Misc} discusses various practical issues, limitations and future research topics regarding the proposed method. Finally, Section \ref{Conclusion} presents the conclusion.

\section{The basic framework of the global AL-GP metamodelling method}\label{Framework}

\subsection{Gaussian process model}

\noindent To make this paper self-contained, the essential concept of Gaussian process model will be briefly reviewed. In Gaussian process metamodelling, the metamodel of $\mathcal{M}(\bm{x})$, denoted by $\hat{\mathcal{M}}(\bm{x})$, is considered as a realization of a Gaussian process \citep{Murphy12}, i.e.

\begin{equation}
\hat{\mathcal{M}}(\bm{x}) \sim GP\left(\mu(\bm{x}),\kappa(\bm{x},\bm{x}')|\bm{\theta}\right)\,,\label{3}
\end{equation}

\noindent where

\begin{equation}
\mu(\bm{x})={\rm E}\left[\hat{\mathcal{M}}(\bm{x})\right]\,\label{4}
\end{equation}

\noindent is the mean function, and

\begin{equation}
\kappa(\bm{x},\bm{x}')={\rm E}\left[\left(\hat{\mathcal{M}}(\bm{x})-\mu(\bm{x})\right)\left(\hat{\mathcal{M}}(\bm{x}')-\mu(\bm{x}')\right)\right]\,\label{5}
\end{equation}

\noindent is a positive definitive kernel function (or covariance function), and $\bm{\theta}$ is a set of parameters describing both the mean and the kernel functions.

Given a training set $\left\lbrace\bm{\mathcal{X}},\bm{\mathcal{Y}}\right\rbrace=\left\lbrace\left(\bm{x}_i,\mathcal{M}(\bm{x}_i)\right),i=1,2,...,d\right\rbrace$, the parameters $\bm{\theta}$ are typically estimated by generalized least-squares solution \citep{Murphy12}. Provided a test set $\left\lbrace\bm{\mathcal{X}}_*,\bm{\mathcal{Y}}_*\right\rbrace=\left\lbrace(\bm{x}_i,\hat{\mathcal{M}}(\bm{x}_i)),i=1,2,...,s\right\rbrace$ for which the predictions $\bm{\mathcal{Y}}_*$ are desired, by definition of the Gaussian process the following holds,

\begin{equation}
\left(
  \begin{array}{ccc}
    \bm{\mathcal{Y}} \\
    \bm{\mathcal{Y}}_*
  \end{array}
\right)\sim\mathcal{N}
\left[
\left(
\begin{array}{ccc}
    \bm{\mu} \\
    \bm{\mu}_*
  \end{array}
\right),
\left(
\begin{array}{ccc}
    \bm{K} & \bm{K}_*\\
    \bm{K}_*^T & \bm{K}_{**}
  \end{array}
\right)
\right]
 \,,\label{6}
\end{equation}

\noindent where $\mathcal{N}$ denotes joint Gaussian distribution, $\bm{\mu}=\mu(\bm{\mathcal{X}})$, $\bm{\mu}_*=\mu(\bm{\mathcal{X}}_*)$, $\bm{K}=\kappa(\bm{\mathcal{X}},\bm{\mathcal{X}})$, $\bm{K}_*=\kappa(\bm{\mathcal{X}},\bm{\mathcal{X}}_*)$, and $\bm{K}_{**}=\kappa(\bm{\mathcal{X}}_*,\bm{\mathcal{X}}_*)$. Using Eq.\eqref{6}, the conditional distribution (or posterior distribution) $f_{\bm{\mathcal{Y}}_*}(\bm{\mathcal{Y}}_*|\bm{\mathcal{Y}})$ can be obtained as

\begin{equation}
\begin{array}{lr}
f_{\bm{\mathcal{Y}}_*}(\bm{\mathcal{Y}}_*\vert\bm{\mathcal{Y}})=\mathcal{N}\left(\bm{\mathcal{Y}}_*\vert\bm{\mu}_{\bm{\mathcal{Y}}_*},\bm{K}_{\bm{\mathcal{Y}}_*}\right)\\
\bm{\mu}_{\bm{\mathcal{Y}}_*}=\bm{\mu}_*+\bm{K}_*^T\bm{K}_*^{-1}\left(\bm{\mathcal{Y}}-\bm{\mu}\right)\\
\bm{K}_{\bm{\mathcal{Y}}_*}=\bm{K}_{**}-\bm{K}_*^T\bm{K}_*^{-1}\bm{K}_*\,\label{7}
\end{array}
\end{equation}

There are several open source toolboxes for training Gaussian process models, e.g. DACE \cite{DACE}, ooDACE \cite{ooDACE}, GPML \cite{GPML}, etc. In this study, the ooDACE package is used to produce the results in Section \ref{Numerical}.

\subsection{Procedures of the global AL-GP metamodelling method}\label{Procedure}

\noindent The basic procedures of the AL-GP approach for distribution function estimation are listed as follows.

\begin{breakablealgorithm}
\label{alg:01}
\caption{Procedures of global AL-GP metamodelling to estimate the CDF $F_Y(y)$}
\begin{description}
\item [Step 1: Initialization]
~\\
\begin{itemize}
\item Generate a large set of training candidates $\bm{\mathcal{X}}_c=\left\lbrace\bm{x}_i,i=1,2,...,N\right\rbrace$.
\item Generate an initial training set $\bm{\mathcal{X}}=\left\lbrace\bm{x}_i,i=1,2,...,d\right\rbrace$, $d<<N$.
\item Evaluate the true model function for $\bm{\mathcal{X}}$ to obtain $\bm{\mathcal{Y}}=\left\lbrace \mathcal{M}(\bm{\mathcal{X}})\right\rbrace$.
\end{itemize}

\item [Step 2: Train the Gaussian process model]
~\\
\begin{itemize}
\item Using $\left\lbrace\bm{\mathcal{X}},\bm{\mathcal{Y}}\right\rbrace$, train a Gaussian process metamodel $\hat{\mathcal{M}}(\bm{x})$.
\end{itemize}
\item [Step 3: Monte Carlo simulation on the metamodel]
~\\
\begin{itemize}

\item Perform Monte Carlo simulation with $\hat{\mathcal{M}}(\bm{x})$ and $\bm{y}$ to obtain a three-fold estimate of $F_Y(y)$, denoted by $\hat{F}_Y^+(\bm{y})$, $\hat{F}_Y^0(\bm{y})$, $\hat{F}_Y^-(\bm{y})$, and $\forall y_i\in\bm{y}$, $\hat{F}_Y^+({y}_i)\ge \hat{F}_Y^0({y}_i)\ge \hat{F}_Y^-({y}_i)$.
\end{itemize}

\item [Step 4: Stopping criterion check]
~\\
\begin{itemize}
\item Evaluate an error measure using $\hat{F}_Y^+(\bm{y})$, $\hat{F}_Y^0(\bm{y})$, and $\hat{F}_Y^-(\bm{y})$.
\item If a specified stopping criterion is met, terminate the algorithm; else, proceed to \textbf{Step 5}.
\end{itemize}

\item [Step 5: Update the training data set]
~\\
\begin{itemize}
\item Search in $\bm{\mathcal{X}}_c$ for the optimal sample that maximizes a specified learning function.
\item Add the optimal sample to the training set $\bm{\mathcal{X}}$.
\item Evaluate the true model function for the optimal sample and update $\bm{\mathcal{Y}}$.
\item Generate a new set of training candidates $\bm{\mathcal{X}}_c$.
\item Return to \textbf{Step 2}.
\end{itemize}
\end{description}
\end{breakablealgorithm}
To obtain the CCDF, we simply use $\bar F_Y(y)=1-F_Y(y)$. The error measure and learning function will be specially designed such that $1-F_Y(y)$ will be accurate in the low probability region of $\bar F_Y(y)$.

In this study we develop on Algorithm \ref{alg:01} by  providing a novel AL-GP strategy such that: a) the CDF as well as the CCDF estimation for a relatively wide range of $y$ is accurate, and b) the efficiency of the method is not sensitive to the CDF/CCDF discretization, i.e. it is mesh free. To achieve these goals, in the following section the crucial ingredients of Algorithm \ref{alg:01} will be developed. Note that active learning based metamodellings all share a similar set of procedures as described in Algorithm \ref{alg:01}. The novelty of this study lies in the global learning strategy to obtain the distribution function (CDF as well as CCDF). In particular, the strategy is based on a novel error measure combined together with a two step learning function, which allows a mesh free estimate of the CDF/CCDF.

\section{A detailed development of the global AL-GP metamodelling method}\label{Details}

\subsection{Initialization of training samples (Step 1 of Algorithm \ref{alg:01})}

\noindent The training candidates $\bm{\mathcal{X}}_c$ can be generated via Monte Carlo simulation using the PDF $f_{\bm{X}}(\bm{x})$, and the sample size $N$ is typically of the order of $10^6$ (or larger). The initial training set $\bm{\mathcal{X}}$ can be generated by quasi Monte Carlo or low-discrepancy sequence techniques such as the Latin hypercube sampling and the Sobol sequence, and the sample size $d$ is typically of the order of $10^1$.

\subsection{Monte Carlo simulation for the metamodel (Step 3 of Algorithm \ref{alg:01})}
\noindent In Gaussian process metamodelling, the $\mathcal{M}(\bm{x})$ is considered as a realization of a Gaussian process at location $\bm{x}$, which is completely defined by the mean $\mu_{\hat{\mathcal{M}}}(\bm{x})$ and variance $\sigma^2_{\hat{\mathcal{M}}}(\bm{x})$ (this mean and variance correspond to the posterior distribution described in Eq.\eqref{7}). It follows that a Gaussian process model describes a family of models expressed by
\begin{equation}
\hat{\mathcal{M}}(\bm{x}|k) = \mu_{\hat{\mathcal{M}}}(\bm{x})+ k\sigma_{\hat{\mathcal{M}}}(\bm{x})\,.\label{8}
\end{equation}
For $k=-\bar k, 0, \bar k$, one obtains three metamodels, i.e.
\begin{equation}
\begin{array}{lr}
\hat{\mathcal{M}}^+(\bm{x}) = \mu_{\hat{\mathcal{M}}}(\bm{x})-\bar k\sigma_{\hat{\mathcal{M}}}(\bm{x})\\
\hat{\mathcal{M}}^0(\bm{x}) = \mu_{\hat{\mathcal{M}}}(\bm{x})\\
\hat{\mathcal{M}}^-(\bm{x}) = \mu_{\hat{\mathcal{M}}}(\bm{x})+\bar k\sigma_{\hat{\mathcal{M}}}(\bm{x})\,\label{9}
\end{array}
\end{equation}
\noindent where $\bar k$ is typically fixed around 2. Using Eq.\eqref{CDF} and Eq.\eqref{9}, one obtains a three-fold CDF estimate of $Y$,
\begin{equation}
\hat{F}_Y^a(y)=\int_{\hat{\mathcal{M}}^a\left(\bm{x}\right)\leq y}\ f_{\bm{X}}(\bm{x})\,d\bm{x}\,,\label{10}
\end{equation}
\noindent where $a=+,0,-$. If a crude Monte Carlo simulation is used to solve Eq.\eqref{10}, to make the estimates consistent, the same set of random samples should be used in estimating $\hat{F}_Y^+(y)$, $\hat{F}_Y^0(y)$, and $\hat{F}_Y^-(y)$\footnote{Due to statistical errors of Monte Carlo solutions, if different set of random samples are used, the property $\hat{F}_Y^+(y)\ge \hat{F}_Y^0(y)\ge \hat{F}_Y^-(y)$ may not hold. This is what we meant by ``consistent''.}.

\subsection{The error measure (Step 4 of Algorithm \ref{alg:01})}
\noindent
The error measure is the critical ingredient for meeting the goal of simultaneously estimating the CDF and CCDF.
To satisfy this goal, the error measure needs to be defined such that the CDF and the CCDF errors are measured \emph{symmetrically}.
Formally, consider a functional $D(\bm F)$, where $\bm F$ denotes a set of CDFs or CCDFs with their discrepancy being measured by $D(\cdot)$. In this study, $\bm F$ may include $\hat{F}_Y^+(y)$, $\hat{F}_Y^0(y)$, and $\hat{F}_Y^-(y)$. Ideally, $D(\bm F)$ should satisfy the following symmetry,

\begin{equation}
D(\bm F)=D(\bm 1-\bm F)\,.\label{Sym}
\end{equation}
The symmetry indicates that the error measure should be invariant under the transformation $\bm F\leftarrow\bm 1-\bm F$. In other words, the error measure $D(\cdot)$ cannot tell if it is the CDF or the CCDF being measured.
Clearly, in Algorithm \ref{alg:01}, the computations should stop if $\hat{F}_Y^+(y)$ is sufficiently close to $\hat{F}_Y^-(y)$. To measure the discrepancy between CDFs $\hat{F}_Y^+(y)$ and $\hat{F}_Y^-(y)$, a natural metric that satisfies Eq.\eqref{Sym} is the Wasserstein (Kantorovichâ-Rubinstein) distance \cite{Wasserstein}, expressed by
\begin{equation}
W(\hat{F}_Y^+,\hat{F}_Y^-)=\int_{-\infty}^{+\infty}\left\vert\hat{F}_Y^+(y)-\hat{F}_Y^-(y)\right\vert\,dy\,.\label{11}
\end{equation}
\noindent Eq.\eqref{11} is not the definition of the Wasserstein metric, but for the unidimensional case one can show that Eq.\eqref{11} is equivalent to the Wasserstein metric \cite{Wasserstein}. The Wasserstein metric corresponds to the minimum ``cost" of turning one distribution into another. In this context, the cost is defined as the amount of probability mass needed to be transported integrated over the ``transportation distance.'' The absolute function in Eq.\eqref{11} seems redundant because by definition $\hat{F}_Y^+(y)$ must be larger or equal to $\hat{F}_Y^-(y)$. However, the absolute function cannot be deleted because otherwise the symmetric condition (Eq.\eqref{Sym}) would be violated. Specifically, the absolute function is reserved for the scenario such that one replaces $\hat{F}_Y^+(y)$ and $\hat{F}_Y^-(y)$ with $1-\hat{F}_Y^+(y)$ and $1-\hat{F}_Y^-(y)$, respectively, and in this scenario Eq.\eqref{11} should provide the identical answer.

The problem of using Eq.\eqref{11} is that the contribution from the distribution tail is negligible. To highlight the tail contribution, we introduce the following symmetric measure
%
%
%
\begin{equation}
W^*(\hat{F}_Y^+,\hat{F}_Y^0,\hat{F}_Y^-)=\int_{-\infty}^{+\infty}w^*(y)dy.\label{EM}
\end{equation}
where

\begin{equation}
w^*(y)\equiv\frac{\left\vert\hat{F}_Y^+(y)-\hat{F}_Y^-(y)\right\vert}{\min\left[\hat{F}_Y^0(y),1-\hat{F}_Y^0(y)\right]}\,.\label{Integrand}
\end{equation}
It is easy to verify that Eq.\eqref{EM} satisfies Eq.\eqref{Sym}. In practice, we replace the infinite integral bound to $[y_{\min},y_{\max}]$, i.e. the range of practical interest, and thus the potential zero denominator issue in Eq.\eqref{EM} can be avoided. With Eq.\eqref{EM}, the stopping criterion can be specified as

\begin{equation}
W^*(\hat{F}_Y^+,\hat{F}_Y^0,\hat{F}_Y^-)<\epsilon\,.\label{StopC}
\end{equation}

\noindent The threshold $\epsilon$ can be set to

\begin{equation}
\epsilon=\bar \epsilon(y_{\max}-y_{\min})\,,\label{epsilon}
\end{equation}

\noindent where $\bar \epsilon$ is a specified tolerance. The tolerance $\bar \epsilon$ indicates that the integrand of Eq.\eqref{EM}, $w^*(y)$, on average (with respect to $y$) should be smaller than $\bar \epsilon$.

\subsection{The learning function (Step 5 of Algorithm \ref{alg:01})}

\noindent The learning function predicts the value of information gained by adding a new sample to the training set. In the AL-GP approach for reliability analysis, the learning function is defined as the probability of misclassification (failure/safe domain) \cite{echard2011ak}\cite{schobi2016rare}, i.e.

\begin{equation}
\bm{L}(\bm{x}|y)=\Phi\left[-\frac{\vert y-\hat{\mathcal{M}}^0(\bm{x})\vert}{\sigma_{\hat{\mathcal{M}}}(\bm{x})}\right]\,,\label{13}
\end{equation}

\noindent where $\Phi[\cdot]$ denotes the CDF function of the univariate standard Gaussian distribution. The learning function Eq.\eqref{13} implies that one should select the optimal training sample, $\bm{x}^*$, such that: a) $\bm{x}^*$ is close to the limit-state surface $y-\hat{\mathcal{M}}^0(\bm{x})=0$, and b) the metamodelling uncertainty, $\sigma_{\hat{\mathcal{M}}}$, at $\bm{x}^*$ is large.

The learning function Eq.\eqref{13} cannot be used in the global AL-GP method because for CDF/CCDF estimation $y$ is not fixed, in other words, for any candidate $\bm{x}$ there will be a corresponding $y$ to exactly have $y-\hat{\mathcal{M}}^0(\bm x)=0$. A simple remedy to this issue is to use the maximum of variance (MoV) learning criterion, i.e.

\begin{equation}
\bm{L}_{MoV}(\bm{x})=\sigma_{\hat{\mathcal{M}}}(\bm{x})\,.\label{14}
\end{equation}

The MoV learning function may not be most effective because it only uses a small portion of the available information and it is not directly related to the error measure Eq.\eqref{EM}. In the following, we will develop an alternative learning function such that: a) it makes full use of the available information, and b) it is directly related to the error measure.
We first apply a kernel to Eq.\eqref{EM} to obtain a localized error measure for the distribution function, i.e.
\begin{equation}
W^*_L(y')=\int_{-\infty}^{+\infty}w^*(y)\psi(y|y',\bm{\nu})\,dy=\int_{-\infty}^{+\infty}\frac{\left\vert\hat{F}_Y^+(y)-\hat{F}_Y^-(y)\right\vert}{\min\left[\hat{F}_Y^0(y),1-\hat{F}_Y^0(y)\right]}\psi(y|y',\bm{\nu})\,dy\,,\label{15}
\end{equation}
where $\psi(y|y',\bm{\nu})$, $\psi(y|y',\bm{\nu})\geq 0$, $\int_{-\infty}^{+\infty}\psi(y|y',\bm{\nu})\,dy=1$, denotes the kernel function centered at $y'$ and parametrized by $\bm{\nu}$. Eq.\eqref{15} measures the localized contribution to the CDF/CCDF error in the neighbourhood of $y'$. The learning function is used to predict the specific ``location'' to assign the training sample, therefore it must contain some ``localized'' information on the error measure. At the same time, the learning function should be provided with some ``global'' knowledge, since the ultimate goal of learning is to reduce a global error measure on the distribution function. Clearly, Eq.\eqref{15} bridges the gap between global and local error measures on the distribution function, and therefore being an ideal medium for constructing a learning function. Given that, we introduce a two-step learning criterion. Specifically, first,  we select the candidate threshold, $y^*$, which is located at the region with maximum localized distribution function error; second, we select the  training sample, $\bm{x}^*$, where the model uncertainty is large. Formally, the two-step learning function is defined as
\begin{equation}
\bm{L}(\bm{x})=\Phi\left[-\frac{\vert y^*-\hat{\mathcal{M}}^0(\bm{x})\vert}{\sigma_{\hat{\mathcal{M}}}(\bm{x})}\right]\,,\label{16}
\end{equation}
\noindent where
\begin{equation}
y^*=\mathop{\arg\max}_{y'} \left\lbrace W^*_L(y')\vert y'\in[y_{\min},y_{\max}]\right\rbrace\,.\label{17}
\end{equation}
Note that the constraint in Eq.\eqref{17} is applied because in practice we are only interested in the CDF/CCDF within the range $[y_{\min},y_{\max}]$.
With the learning function, the optimal training sample is selected as

\begin{equation}
\bm{x}^*=\mathop{\arg\max}_{\bm x\in\bm{\mathcal{X}}_c}\left\lbrace\bm{L}(\bm{x})\bigg\vert \hat{\mathcal{M}}^0(\bm x)\in \left[y_{\min}-\bar k\sigma_{\hat{\mathcal{M}}}(\bm x),y_{\max}+\bar k\sigma_{\hat{\mathcal{M}}}(\bm x)\right]\right\rbrace\,.\label{18}
\end{equation}
With this strategy, for a given training candidate set $\mathcal{X}_c$ $\bm{x}^*$ is the point associated with the maximum localized distribution function error and---at the same time---with the largest model uncertainty. Similar to Eq.\eqref{17}, the constraint in Eq.\eqref{18} is applied because $[y_{\min},y_{\max}]$ is of interest, and the term $\bar k\sigma_{\hat{\mathcal{M}}}(\bm x)$ is used to consider the model uncertainty of $\hat{\mathcal{M}}(\bm x)$.
\subsubsection{Kernel selection: Dirac kernel}
\noindent
A special case of Eq.\eqref{15} is the use of Dirac kernel $\delta(y-y')$, and Eq.\eqref{15} reduces to
\begin{equation}
W^*_L(y')=\int_{-\infty}^{+\infty}w^*(y)\delta(y-y')\,dy=w^*(y)\,.\label{19}
\end{equation}
The use of Dirac Kernel corresponds to a greedy approach, and for this case the optimization described by Eq.\eqref{17} is trivial.
\vspace{-2mm}
\subsubsection{Kernel selection: Gaussian kernel}
\begin{figure}[h]
  \centering
  \includegraphics[scale=0.5]{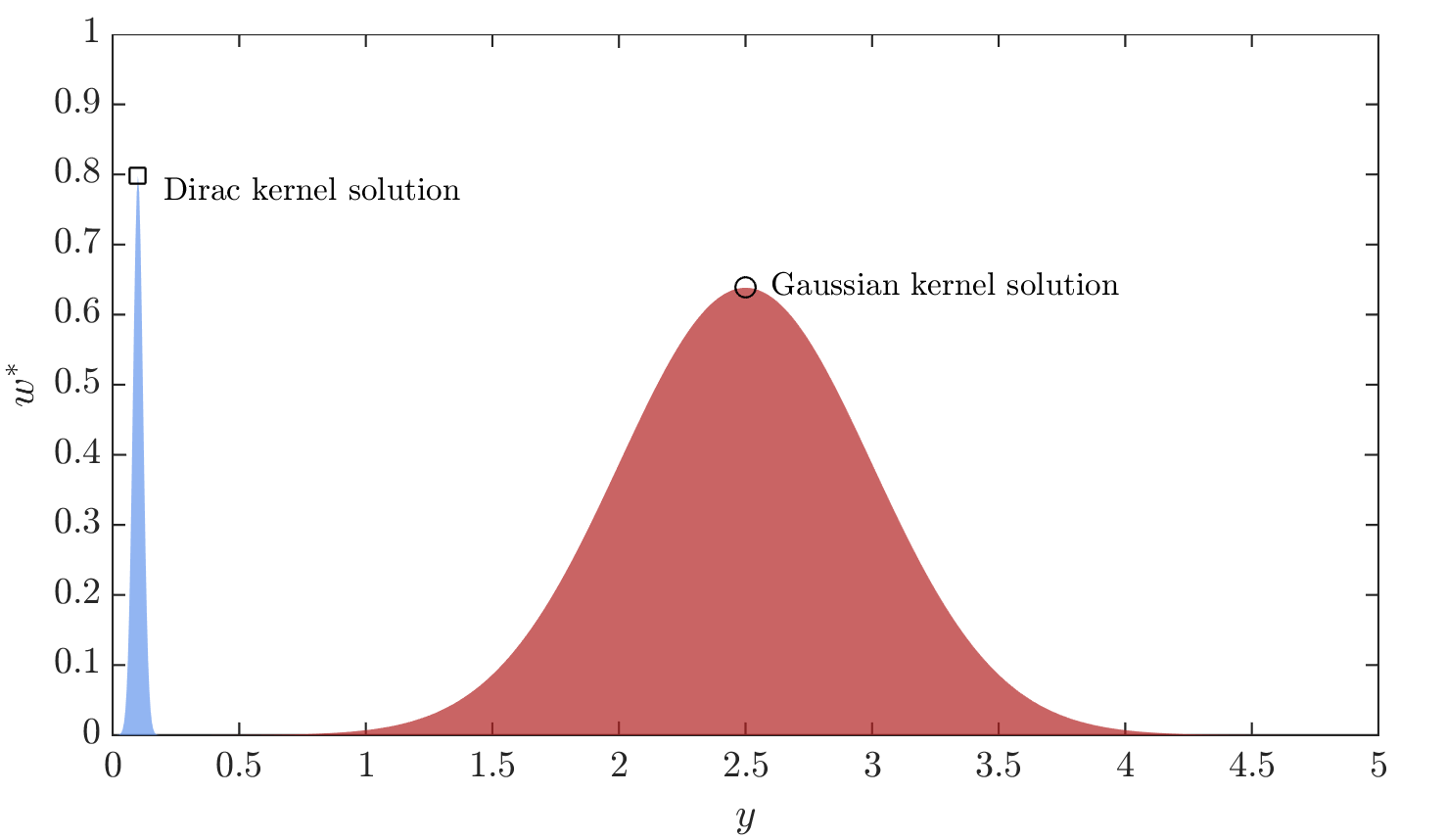}
  \caption{\textbf{Illustration of the effects of kernel function}. \emph{The figure shows the integrand of Eq.\eqref{EM}, $w^*(y)$, as a function of $y$. If a Dirac kernel is used, the next training candidate will be put on the left peak (the maximum). (Ignoring, for the moment, the influence of model uncertainty $\sigma_{\mathcal{\hat M}}$.) This threshold candidate may not be as effective as putting on the middle peak to reduce the global error measure $W^*$, since putting a training candidate on the middle peak will have the benefit to suppress the widely distributed error in the neighborhood regions}.}\label{Fig1}
\end{figure}

\noindent
If a Gaussian kernel is used, Eq.\eqref{15} can be written as

\begin{equation}
W^*_L(y')=\frac{1}{\sqrt{2\pi\sigma^2}}\int_{-\infty}^{+\infty}w^*(y)e^{-\frac{(y-y')^2}{2\sigma^2}}\,dy\,.\label{20}
\end{equation}
The standard deviation $\sigma$ controls the width of the kernel. The limiting case $\sigma\to 0$ corresponds to the Dirac kernel, and $\sigma\to +\infty$ corresponds to a uniform distribution. Setting $\sigma\to +\infty$ is clearly not desirable, because this implies an almost constant $y^*$ solution for Eq.\eqref{17}. Setting $\sigma\to 0$, i.e. the Dirac kernel, may also be undesirable, because it does not consider the possible contribution to the error measure $W^*$ from the neighborhood of $y^*$. Figure \ref{Fig1} illustrates this concept. In terms of Figure \ref{Fig1}, if $\sigma$ is properly selected, the next training point would be put on the ideal location which properly accounts for the possible contribution from the vicinity of $y^*$, thus leading to an effective route to reduce the global error measure $W^*$.

With the insight of how $\sigma$ influences the training, we note that $\sigma$ is related to the ``effective bandwidth" of applying a training sample at location $y'$. The information of the effective bandwidth can be extracted from Eq.\eqref{9}. It is seen from Eq.\eqref{9} that applying a training sample at $y'=\hat{\mathcal{M}}^0(\bm{x})$ directly influences probability estimate of $y'\pm\bar k\sigma_{\hat{\mathcal{M}}}(\bm{x})$\footnote{Note that $\mathbb{P}\left(\hat{\mathcal{M}}^{\pm}\leq y'\right)=\mathbb{P}\left(\hat{\mathcal{M}}^0\leq y'\pm\bar k\sigma_{\hat{\mathcal{M}}}(\bm{x})\right)$.}; thus, one can simply set the effective bandwidth $b_e=\bar k\sigma_{\hat{\mathcal{M}}}(\bm{x})$. Since $\bar k$ is commonly fixed to $2$, and assuming the Gaussian kernel is ``effective" in the $\pm2\sigma$ regions around the mean, we can set $\sigma=\sigma_{\hat{\mathcal{M}}}(\bm{x})$. However, $\sigma_{\hat{\mathcal{M}}}(\bm{x})$ is a function of $\bm x$, while, in the current context, it is ideal to let $\sigma$ be solely determined by $y'$. In principle this is not possible because the effective bandwidth $b_e$ must be a function of $\bm x$. However, observe that in practical implementation we have a finite training candidate set $\mathcal{X}_c$. Therefore, in $\mathcal{X}_c$ we can select the sample in the closest neighborhood of $y'$ and assign the corresponding $\sigma_{\hat{\mathcal{M}}}$ as the $\sigma$ for the Gaussian kernel.


Formally, in the Gaussian kernel approach we solve the following optimization.
\begin{equation}
y^*=\mathop{\arg\max}_{y'}\left\lbrace\frac{1}{Z}\int_{y_{\min}}^{y_{\max}}w^*(y)e^{-\frac{(y-y')^2}{2\sigma^2(y')}}\,dy\bigg\vert y'\in[y_{\min},y_{\max}]\right\rbrace\,,\label{21}
\end{equation}
\noindent where the normalizing constant $Z$ is expressed as
\begin{equation}
Z=\sqrt{2\pi}\sigma(y')\left(\Phi\left(\frac{y_{\max}-y'}{\sigma(y')}\right)-\Phi\left(\frac{y_{\min}-y'}{\sigma(y')}\right)\right)\,,\label{22}
\end{equation}
and $\sigma(y')$ is solved from
\begin{equation}
\label{sigma}
  \sigma(y')=\left\lbrace\sigma_{\hat{\mathcal{M}}}(\bm x')\Big\vert\bm x'=\mathop{\arg\min}_{\bm x\in\mathcal{X}_c}\big\vert y'-\hat{\mathcal{M}}^0(\bm x)\big\vert\right\rbrace
\end{equation}
Note that Eq.\eqref{sigma} is adopted for simplicity, yet one could develop other approaches via interpolation or the k-Nearest Neighbor. Also note that the introduction of a correction in the normalizing constant of Gaussian distribution is because the Gaussian kernel is truncated by the range $[y_{\min},y_{\max}]$.

In the global AL-GP approach with Gaussian kernel there is no parameter to tune. Although analytical solution of Eq.\eqref{21} cannot be developed, the one dimensional constrained optimization problem can be effectively solved by various global optimization algorithms (e.g. simulated annealing \cite{SimA} and differential evolution \cite{DifE}).
We finally remark that the error measure as well as the learning procedure are constructed based on a global description of the CDF/CCDF, thus naturally the proposed AL-GP method is mesh free (no discretization of CDF or CCDF are introduced a-priori). Moreover, the proposed error measure is equipped with the important CDF-CCDF symmetry; as a consequence, the proposed AL-GP strategy is designed to accurately estimate the CDF and CCDF at the same time. 

\section{Numerical examples}\label{Numerical}

\subsection{A toy example}

\noindent Consider an analytical two dimensional model function $\mathcal{M}(\bm X)$ expressed by

\begin{equation}
Y=\mathcal{M}(\bm X)=\min\left[X_1-X_2,X_1+X_2\right]\,,\label{23}
\end{equation}

\noindent where $X_1$ and $X_2$ are independent standard Gaussian random variables. The CDF of $Y$ can be expressed by

\begin{equation}
F_Y(y)=\Phi\left(\frac{y}{\sqrt{2}}\right)\left(2-\Phi\left(\frac{y}{\sqrt{2}}\right)\right)\,.\label{24}
\end{equation}
Although Eq.\eqref{24} is not analytical, most scientific computing codes provide highly accurate built-in univariate standard Gaussian CDF $\Phi(\cdot)$, and we denote the corresponding solution as ``exact."
Then, the proposed global AL-GP approaches with the Dirac and Gaussian kernels in Eq.\eqref{15}, and the approach using the maximum of variance (MoV) learning criterion\footnote{The approach using the MoV learning function is also viewed as a global AL-GP method, because it also uses the global error measure Eq.\eqref{EM}.} are used to estimate the distribution function. The conventional AL-GP approach\footnote{In the conventional AL-GP approach, we start from estimating the probability for a fixed threshold with learning function \eqref{13} and stopping criterion $w^*<\bar\epsilon$, and then we iteratively reuse the current metamodel and perform AL-GP to estimate the probability for the next nearby threshold.} is also considered for a comparison. The CDF/CCDF range of interest, $[y_{\min},y_{\max}]$, is set to $[-5,3]$. The range $[y_{\min},y_{\max}]$ is uniformly discretized into $100$ intervals. We use $\epsilon=\bar\epsilon(y_{\max}-y_{\min})$, $\bar\epsilon=0.2$, for the stopping criterion. The number of samples in the training candidate set $\mathcal{X}_c$ is $10^6$. In the following examples except the range $[y_{\min},y_{\max}]$, other settings of the AL-GP method will be the same. Table \ref{tab1ex1} reports the performance of the global AL-GP approaches and the conventional AL-GP approach in estimating the distribution function averaged over $50$ independent runs.

\begin{table}[H]
  \caption{\textbf{Distribution function estimations averaged over $50$ independent runs}. \emph{In the table, ${\rm E}[\epsilon_e]$ denotes expectation of the error measure $\epsilon_e$; $\sigma(\epsilon_e)$ denotes standard deviation of $\epsilon_e$; ${\rm E}[N_{\mathcal{M}}]$ denotes expectation of the number of (true) model function evaluations, and it is composed by the initial training set plus the adaptively added training set}.}
  \label{tab1ex1}
  \centering
  \begin{tabular}{c c c c}
    \toprule
    Methods & ${\rm E}[\epsilon_e]$ & $\sigma(\epsilon_e)$ & ${\rm E}[N_{\mathcal{M}}]$\\
    \midrule
   Gaussian & 0.018 & 0.010 & 12+29.06\\
   Dirac &  0.021 & 0.022 & 12+28.74\\
   MoV & 0.028 & 0.043 & 12+27.34\\
   Conventional & 0.013 & 0.011 & 12+86.22\\
    \bottomrule
  \end{tabular}
\end{table}

\begin{figure}[H]
    \centering
    \includegraphics[scale=0.45]{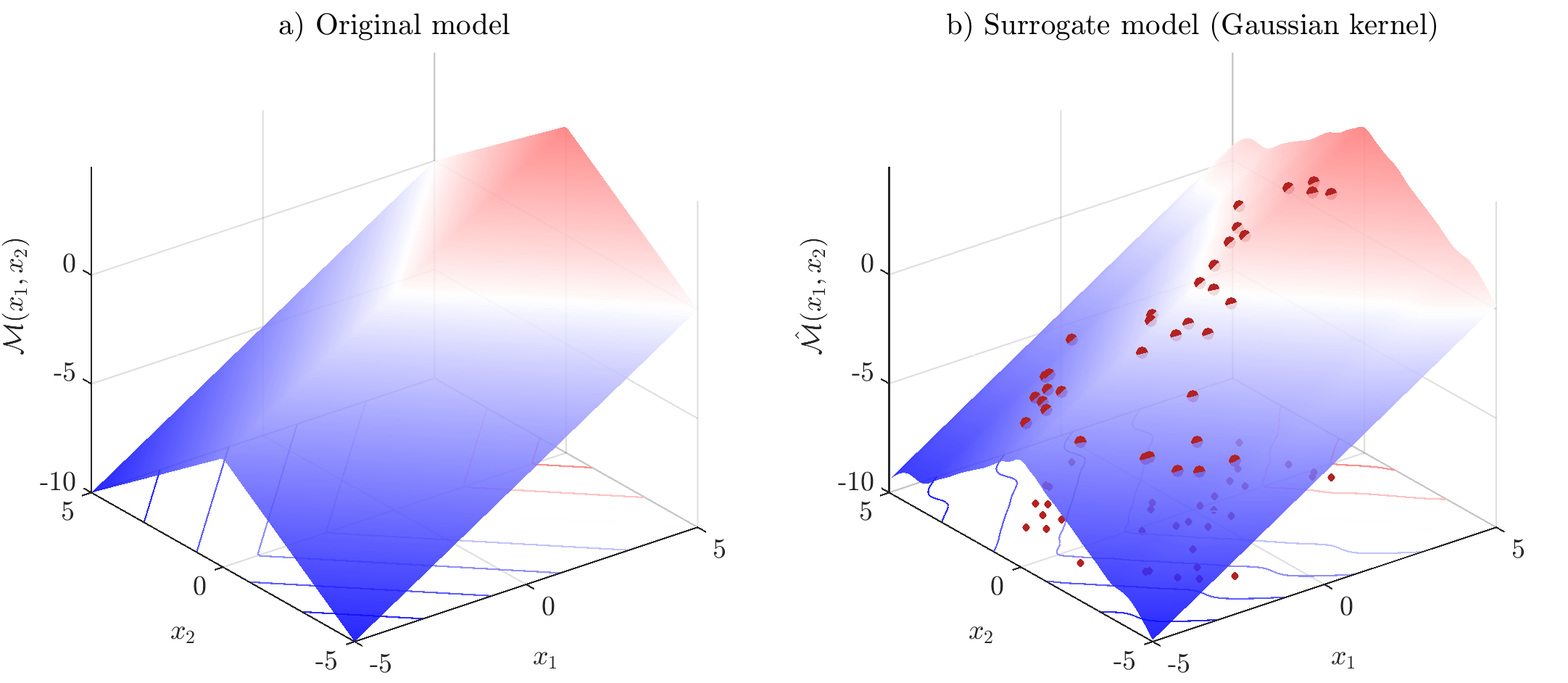}
    \caption{\textbf{A typical Gaussian process metamodel for Eq.\eqref{23}}. a) Original model, b) metamodel based on the red samples (obtained via global AL-GP with Gaussian kernel)
    }
    \label{Fig1Ex1}
\end{figure}

\begin{figure}[H]
    \centering
     \includegraphics[scale=0.45]{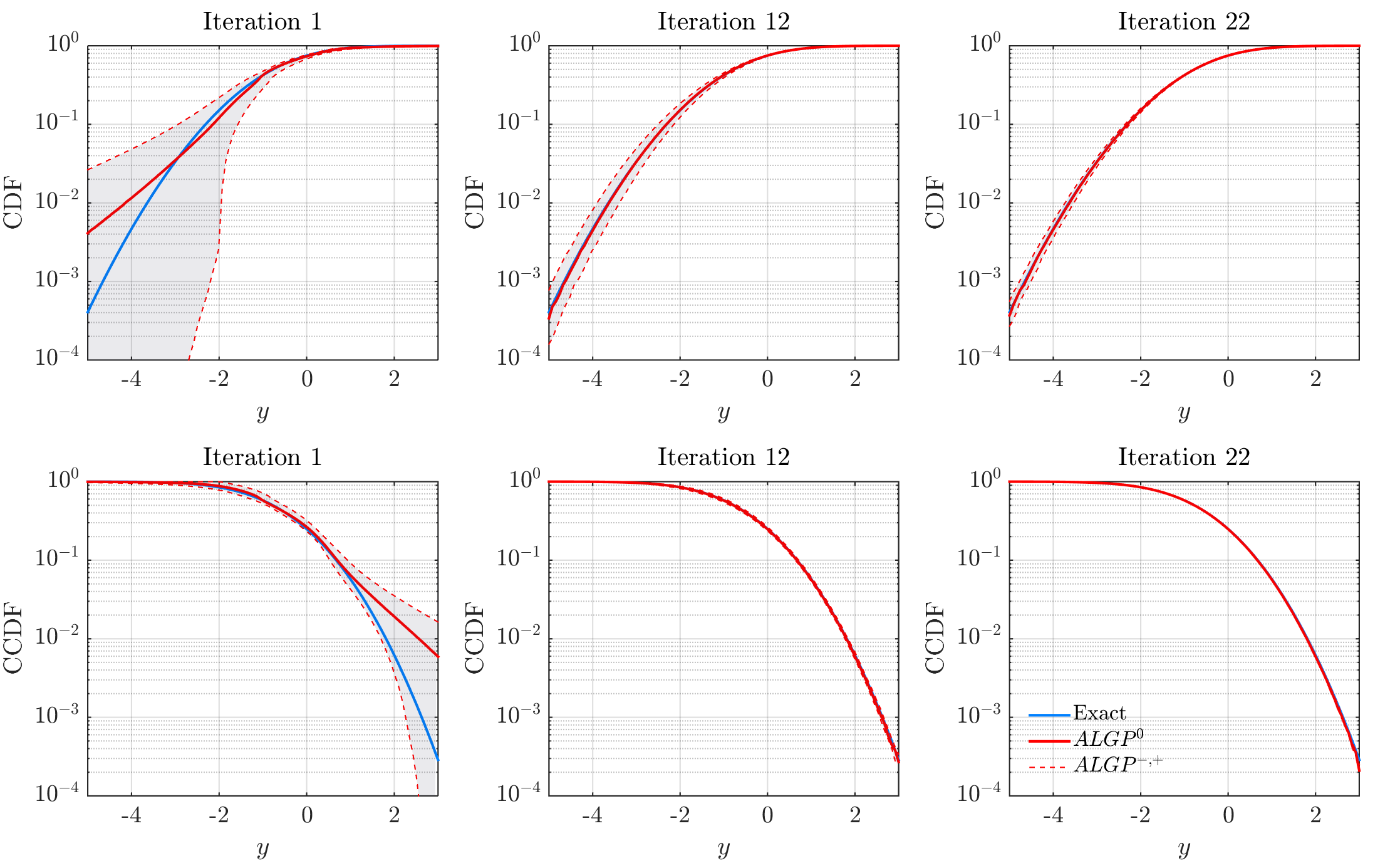}
    \caption{\textbf{Iterative process in distribution function estimation}.}
    \label{Fig2Ex1}
\end{figure}

In the table, the error $\epsilon_e$ is defined as

\begin{equation}
\epsilon_e=\frac{1}{y_{\max}-y_{\min}}\int_{y_{\min}}^{y_{\max}}\frac{\left\vert\hat{F}_Y^0(y)-{F}_Y(y)\right\vert}{\min\left[{F}_Y(y),1-{F}_Y(y)\right]}\,dy\,,\label{25}
\end{equation}

\noindent where ${F}_Y(y)$ is the ``exact" CDF obtained from Eq.\eqref{24}. This error measure also satisfies the CDF-CCDF symmetry condition (Eq.\eqref{Sym}), and it has a clear interpretation. For example, $\epsilon_e=0.02$ indicates, within a specified range of interest, the average (with respect to $y$) relative CDF/CCDF error is $2$ percent. From the table, it can be seen that the global AL-GP approaches are noticeably more efficient than the conventional approach, and the approaches with Gaussian and Dirac kernels are more effective than that with the MoV learning function.
Figure \ref{Fig1Ex1} illustrates a typical Gaussian process metamodel for the model function Eq.\eqref{23}. Figure \ref{Fig2Ex1} illustrates the iterative process for the CDF and CCDF estimation from the global AL-GP approach using the Gaussian kernel.

\begin{figure}[H]
    \centering
    \includegraphics[scale=0.45]{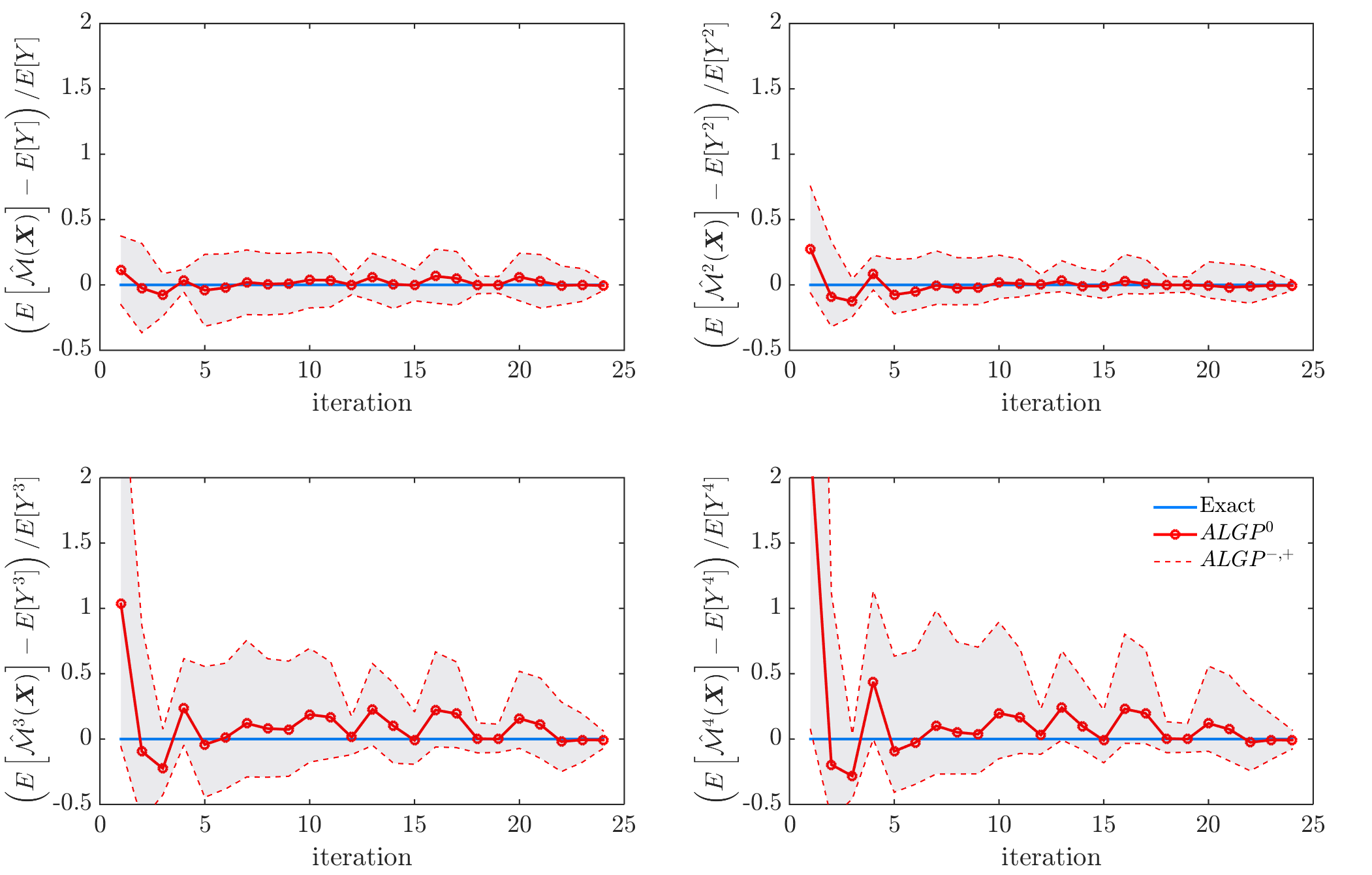}
    \caption{\textbf{Iterative process in moment estimation}.}
    \label{Fig3Ex1}
\end{figure}

\begin{figure}[H]
\centering
  \includegraphics[scale=0.45]{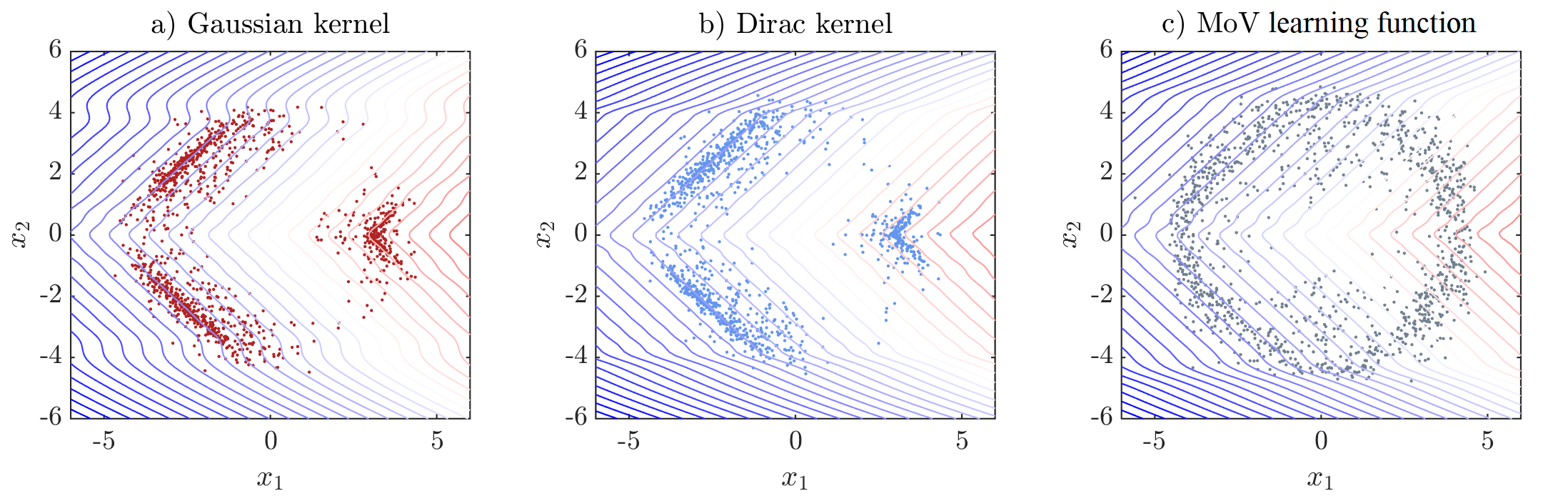}
  \caption{\textbf{Locations of the training samples}.}\label{Fig4Ex1}
\end{figure}

\begin{figure}[H]
\centering
  \includegraphics[scale=0.45]{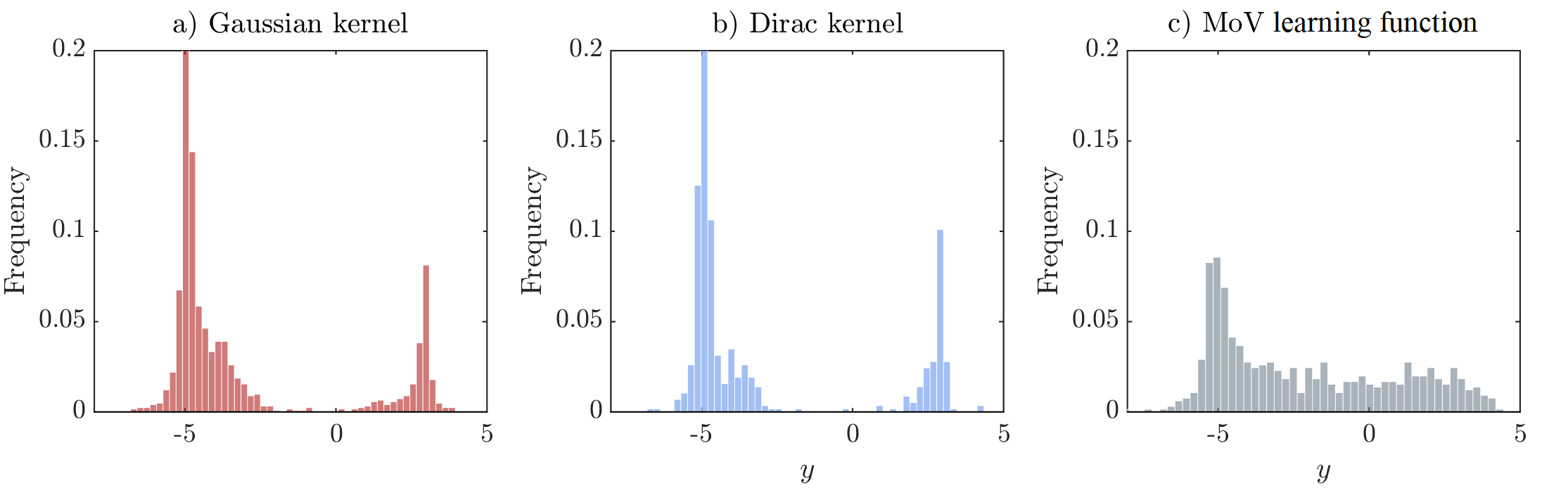}
  \caption{\textbf{Histograms on the $y$ locations of the training samples}.}\label{Fig5Ex1}
\end{figure}

Figure \ref{Fig3Ex1} shows a typical iterative process for the first four statistical moments of the estimated distribution from the global AL-GP approach with Gaussian kernel. The ``exact'' moments are obtained from the following equation,

\begin{equation}
{\rm E}(Y^j)=j\int_{0}^{+\infty}y^{j-1}(1-F_Y(y)+(-1)^jF_Y(-y))\,dy\,.\label{Mom}
\end{equation}
Note that Eq.\eqref{Mom} is in terms of the CDF instead of the PDF, thus being especially useful in this study. Table \ref{tab2ex1} shows the estimated mean, standard deviation, skewness and kurtosis, averaged over $50$ independent runs. The absolute coefficient of variation (c.o.v.) of these estimates are also shown in the table. Note that the number of model function evaluations is already reported in Table \ref{tab1ex1}. Also note that the aforementioned statistical measures can be computed from the first four moments, thus their ``exact'' solutions are obtained via Eq.\eqref{Mom}. It can be seen from the table that the global AL-GP method is able to accurately estimate various statistical measures on the distribution function. More importantly, this accuracy is achieved at the cost of only a few (less than $40$) model function evaluations.
\begin{table}[H]
  \caption{\textbf{Global measures of the estimated distribution function averaged over $50$ independent runs}.}
  \label{tab2ex1}
  \centering
  \begin{tabular}{c c c c c}
    \toprule
    Methods & mean/$|c.o.v.|$ & standard deviation/$|c.o.v.|$& skewness/$|c.o.v.|$ & kurtosis/$|c.o.v.|$\\
    \midrule
   Gaussian & -0.7946/0.0115 & 1.1668/0.0048 &  -0.1391/0.0728 & 3.0619/0.0057\\
   Dirac &  -0.7960/0.0113 & 1.1691/0.0079 & -0.1386/0.1335 & 3.0576/0.0084\\
   MoV &  -0.7996/0.0092 &  1.1700/0.0073 & -0.1312/0.2358 & 3.0538/0.0040\\
   Conventional &  -0.7989/0.0074 &  1.1664/0.0037 & -0.1370/0.0743 & 3.0663/0.0088\\
   Exact & -0.7979 & 1.1676 & -0.1369 & 3.0617\\
    \bottomrule
  \end{tabular}
\end{table}
Figure \ref{Fig4Ex1} shows locations of the training samples and contours of the metamodel. Figure \ref{Fig5Ex1} shows histograms on the $y$ locations of the training samples, i.e. for each training sample $\bm x$, location $y$ is obtained by $y=\mathcal{M}(\bm x)$. Note that Figure \ref{Fig4Ex1} and Figure \ref{Fig5Ex1} illustrate all the adaptively added training samples obtained from $50$ independent runs, thus they depict the spatial distribution of the training samples.
\noindent
It can be seen from Figure \ref{Fig5Ex1} that for the global AL-GP approaches with Dirac or Gaussian kernels, most training samples come from the tails; while for the global AL-GP approach using MoV learning function, training samples are more dispersively distributed. Figure \ref{Fig5Ex1} also suggests that the $y^*$ solution (Eq.\eqref{17}) from the Gaussian kernel, in most cases, is close to the solution from the Dirac kernel. Therefore, the situation discussed in Figure \ref{Fig1} may rarely happen.

\subsection{Ishigami function}

\begin{figure}[H]
    \centering
    \includegraphics[scale=0.45]{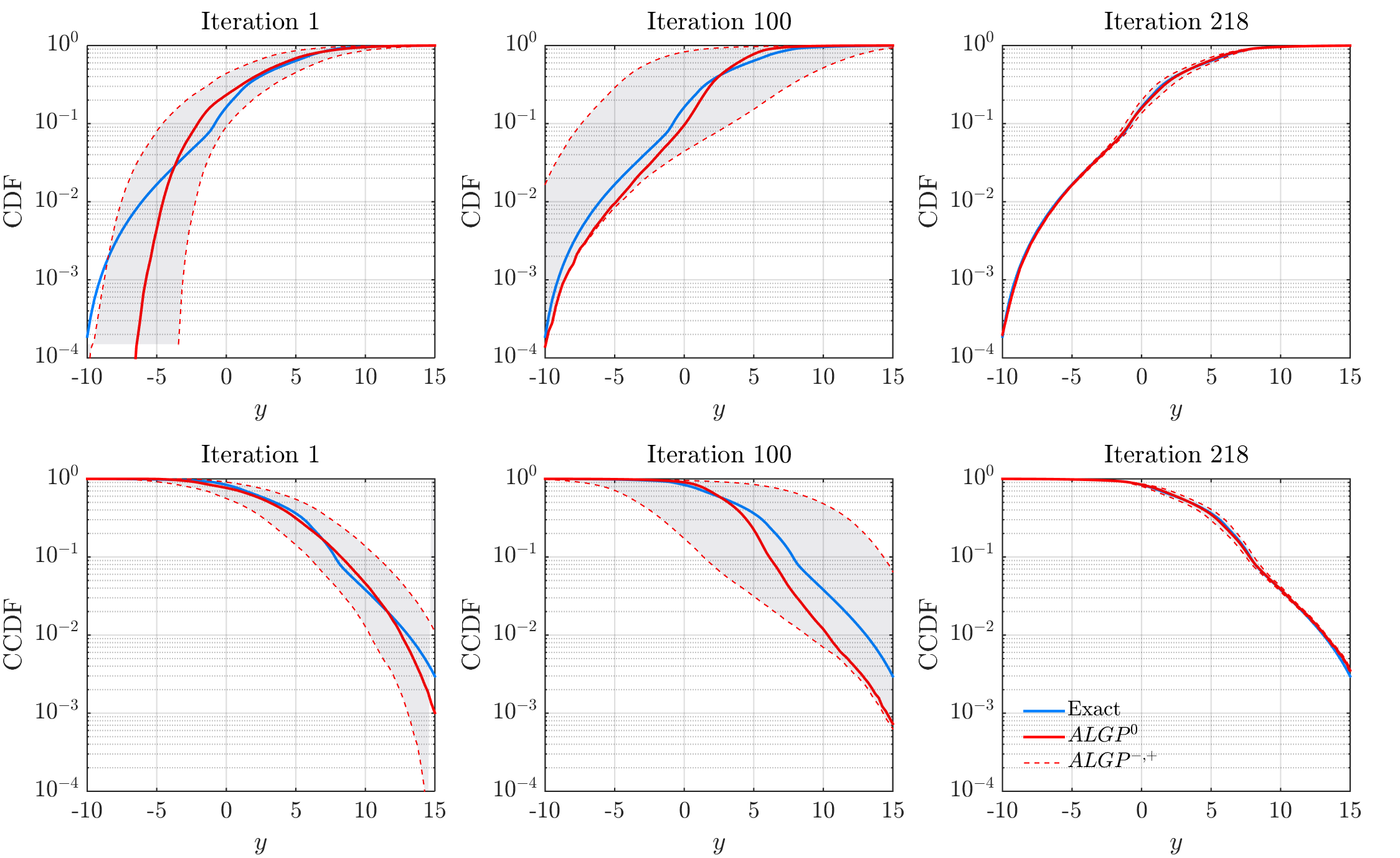}
    \caption{\textbf{Iterative process in distribution function estimation}.}
    \label{Fig1Ex2}
\end{figure}

\begin{figure}[H]
    \centering
    \includegraphics[scale=0.45]{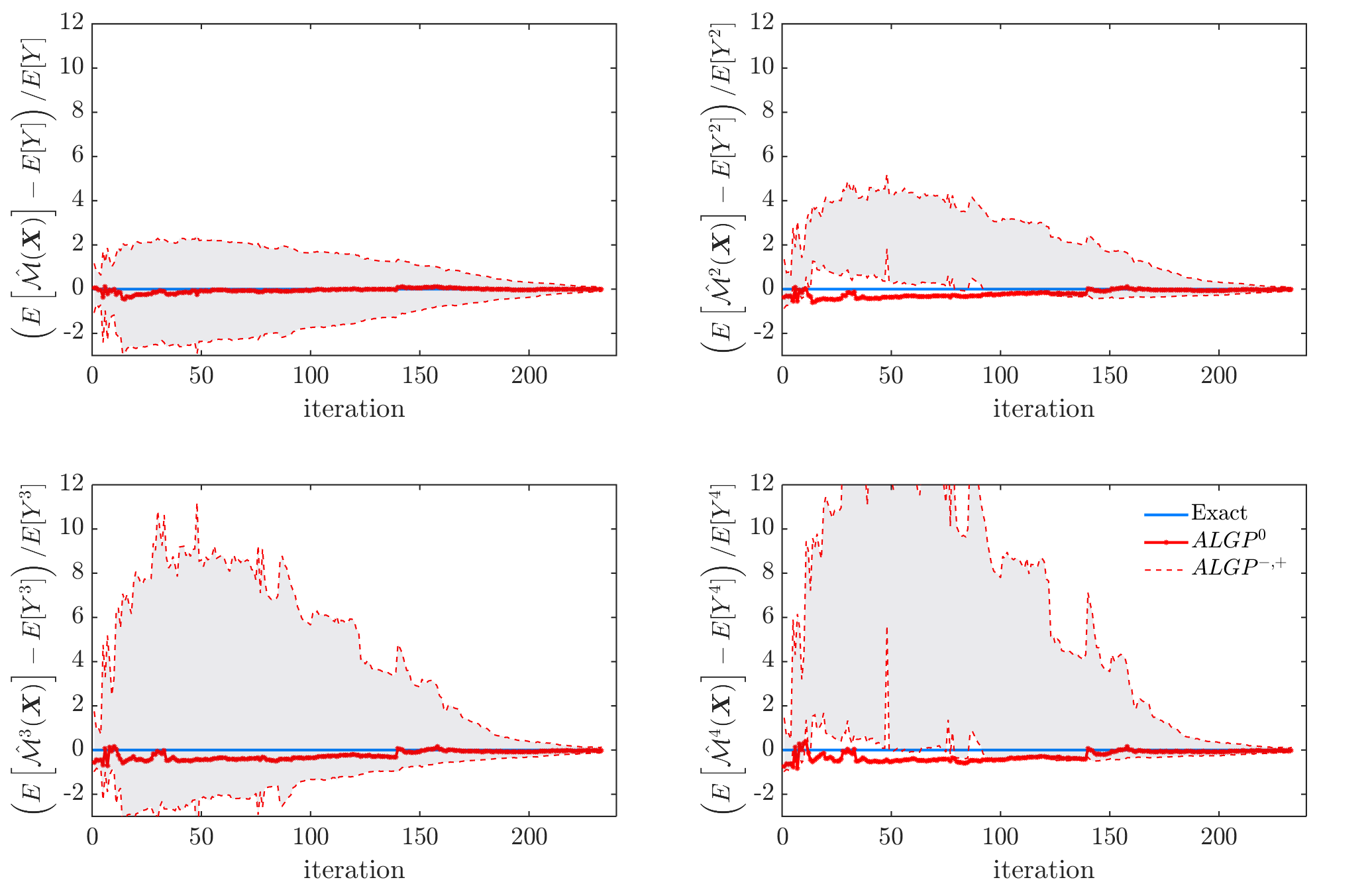}
    \caption{\textbf{Iterative process in moment estimation}.}
    \label{Fig2Ex2}
\end{figure}

\noindent Consider the Ishigami function \cite{ishigami} expressed by

\begin{equation}
Y=\mathcal{M}(\bm X)=\sin(X_1)+a{\sin}^2(X_2)+bX_3^4\sin(X_1)\,,\label{ishi}
\end{equation}

\noindent where $a=7$, $b=0.1$, and $X_1$, $X_2$ and $X_3$ are independent uniform random variables within $[-\pi,\pi]$. The CDF/CCDF range of interest, $[y_{\min},y_{\max}]$, is set to $[-10,15]$. The performance of the global AL-GP approaches are reported in Table \ref{tab1ex2}, and the estimations for the distribution mean, standard deviation, skewness and kurtosis are reported in Table \ref{tab2ex2}. The distribution function estimated via a crude Monte Carlo simulation with $10^7$ samples is used as the reference solution. It is seen from the tables that except the skewness (which is zero theoretically), the global AL-GP approaches provide accurate estimation on the distribution function and various statistical measures. Figure \ref{Fig1Ex2} and Figure \ref{Fig2Ex2} show typical iterative processes for the distribution function and moment estimations from the global AL-GP approach with Gaussian kernel. Observe that for even moments the three fold monotonic property of the CDF is not necessarily preserved (this can be clearly seen from Eq.\eqref{Mom}).

\begin{table}[H]
  \caption{\textbf{Distribution function estimations averaged over 50 independent runs}.}
  \label{tab1ex2}
  \centering
  \begin{tabular}{c c c c}
    \toprule
    Methods & ${\rm E}[\epsilon_e]$ & $\sigma(\epsilon_e)$& ${\rm E}[N_{\mathcal{M}}]$\\
    \midrule
   Gaussian & 0.028 & 0.0074 & 12+230.20\\
   Dirac & 0.027 & 0.0074 & 12+225.60\\
   MoV & 0.060 & 0.0133 & 12+212.20\\
   Conventional & 0.019 & 0.0035 & 12+366.80\\
    \bottomrule
  \end{tabular}
\end{table}

\begin{table}[H]
  \caption{\textbf{Global measures of the estimated distribution function}.}
  \label{tab2ex2}
  \centering
  \begin{tabular}{c c c c c}
    \toprule
    Methods & mean/$|c.o.v.|$ & standard deviation/$|c.o.v.|$& skewness/$|c.o.v.|$ & kurtosis/$|c.o.v.|$\\
    \midrule
   Gaussian & 3.5023/0.0088 & 3.6844/0.0027 & 0.0113/1.6956 & 3.5948/0.0090\\
   Dirac &  3.5204/0.0101 &  3.7062/0.0043 & -0.0036/5.1806 & 3.5603/0.0133\\
   MoV &  3.4933/0.0046 &  3.7789/0.0045 & -0.0010/11.5284 & 3.5087/0.0125\\
   Conventional &  3.5002/0.0026 &  3.7236/0.0018 & 0.0047/2.3023 & 3.5066/0.0054\\
   Monte Carlo & 3.5018  & 3.7204 & 0 (exact)& 3.5106\\
    \bottomrule
  \end{tabular}
\end{table}

\begin{figure}
\centering
  \includegraphics[scale=0.47]{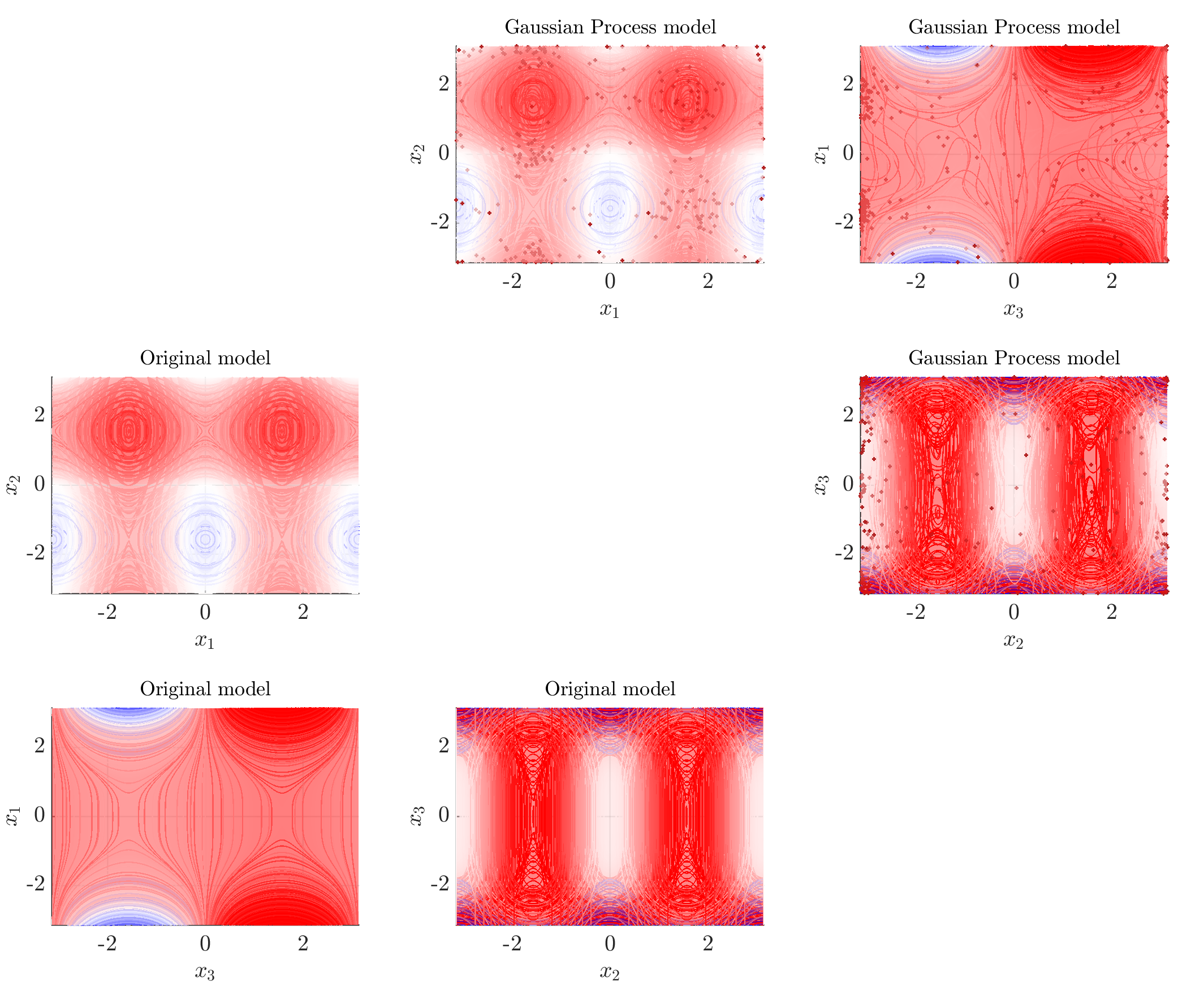}  \caption{\textbf{Projection of the original model and Gaussian process model in various planes}. \textit{The figure shows the projection of level sets $\left\lbrace \bm x\vert m-\mathcal{M}(\bm x)=0\right\rbrace$ and $\left\lbrace \bm x\vert m-\hat{\mathcal{M}}(\bm x)=0\right\rbrace$ in various planes. The training samples used to generate the metamodel is also shown in the figure}.}\label{Fig3Ex2}
\end{figure}

\begin{figure}
\centering
  \includegraphics[scale=0.45]{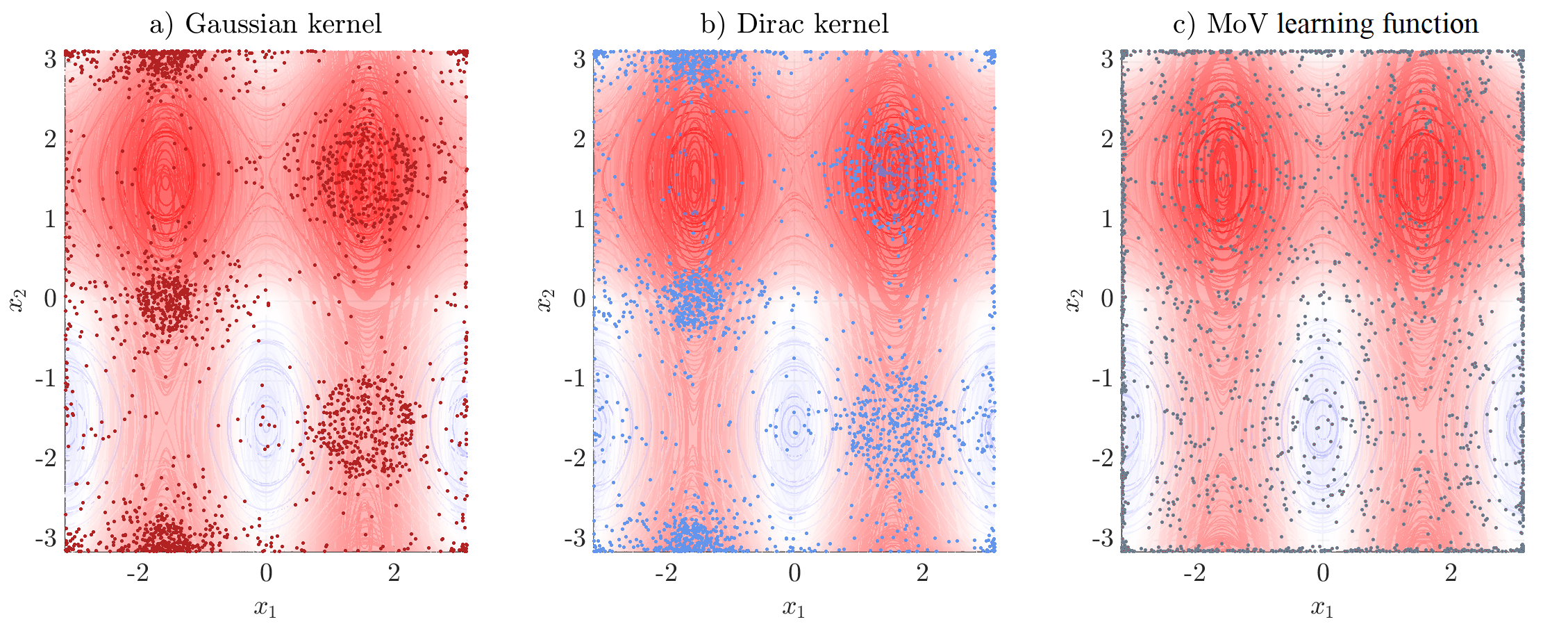}
  \caption{\textbf{Projection of the training samples in the $(x_1,x_2)$ plane}.}\label{Fig4Ex2}
\end{figure}

\begin{figure}
\centering
  \includegraphics[scale=0.45]{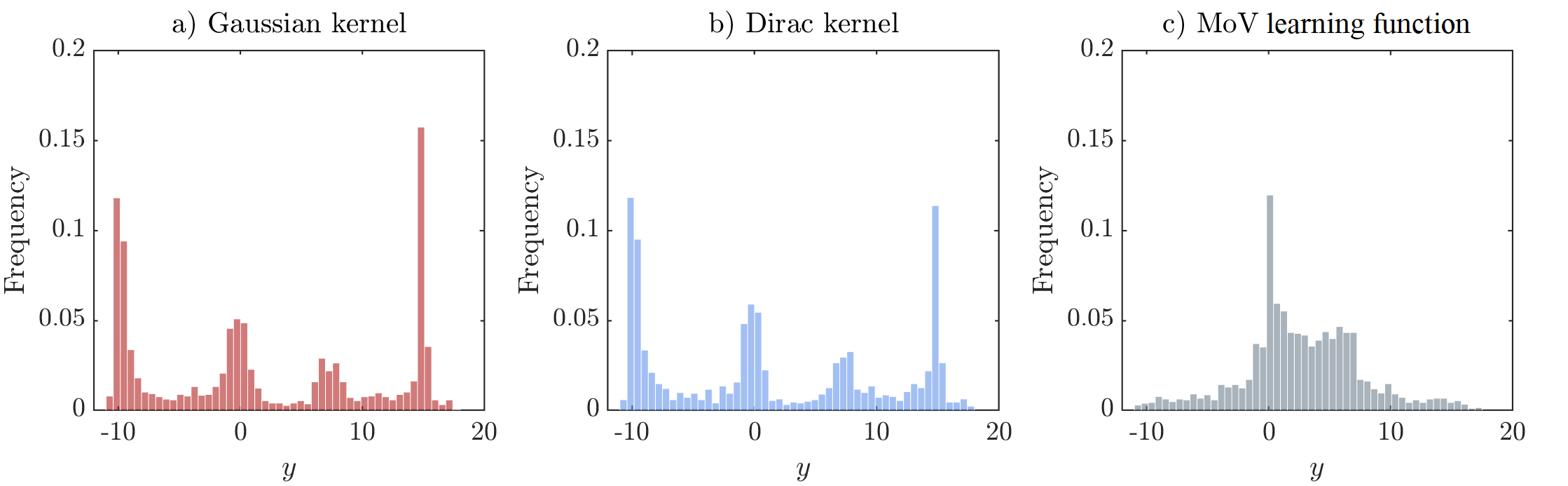}
  \caption{\textbf{Histograms on the $y$ locations of the training samples}.}\label{Fig5Ex2}
\end{figure}

\noindent
Figure \ref{Fig3Ex2} shows projection of the original model and the Gaussian process model in various planes. Figure \ref{Fig4Ex2} shows projection of the training samples in the $(x_1,x_2)$ plane. Figure \ref{Fig5Ex2} shows histograms of the $y$ locations of the training samples. Similar to the previous example, the results obtained from multiple independent runs are used to produce Figure \ref{Fig4Ex2} and Figure \ref{Fig5Ex2}, and only the adaptively added training samples are shown in the figures. Due to the oscillating behavior of the Ishigami function, it is seen from Figure \ref{Fig5Ex2} that the training samples for Gaussian and Dirac kernels concentrate around multiple modes.
\subsection{Structural dynamics analysis of a shear-frame structure}
\begin{figure}[H]
    \centering
    \includegraphics[scale=0.9]{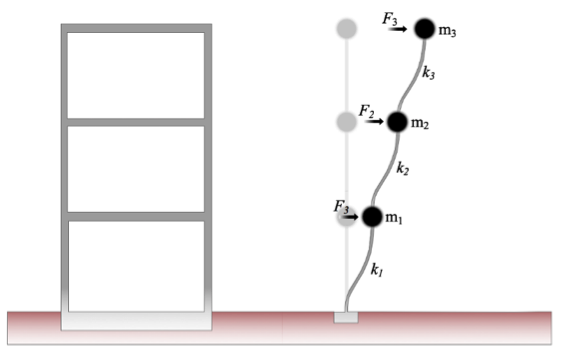}
    \caption{\textbf{Structural archetype}.
    }
    \label{Fig1Ex3}
\end{figure}

\begin{figure}[H]
    \centering
    \includegraphics[scale=0.5]{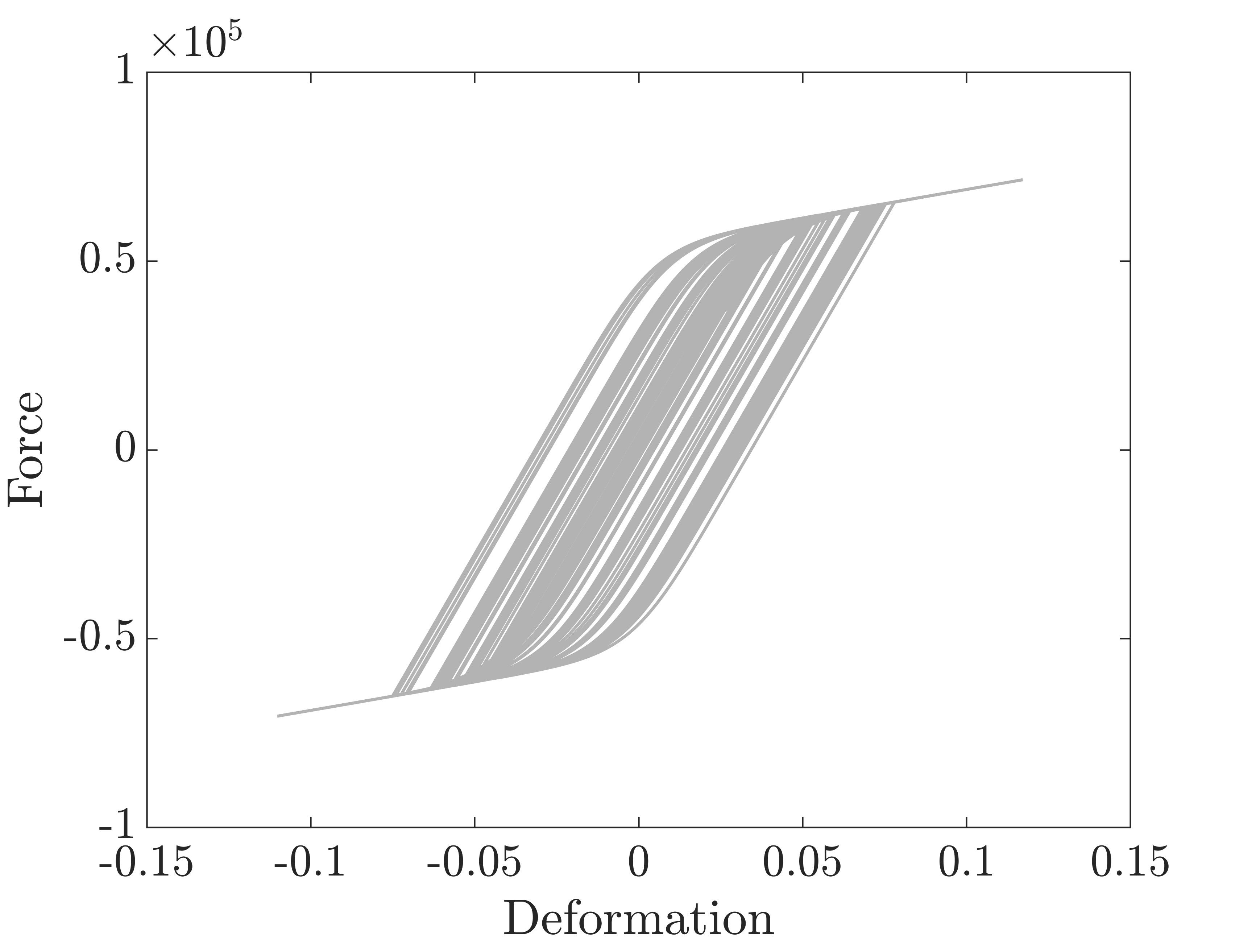}
    \caption{\textbf{Typical hysteretic loops of the Bouc-Wen model}.
    }
    \label{Fig2Ex3}
\end{figure}

\noindent Consider a dynamics analysis of the three stories shear-frame structure shown in Figure \ref{Fig1Ex3} (a similar structural model is studied in \cite{WANG2019}). The interstory mechanical behavior is inelastic with a force-interstory-drift relationship described by the Bouc-Wen hysteretic model \cite{Bouc}\cite{Wen}:
\begin{equation}
\begin{array}{lr}
  k(\alpha u_1(t)+(1-\alpha)u_2(t))=f(t) \\
  \dot{u}_2(t)=-\gamma\left|\dot{u}_1(t)\right|\left|u_2(t)\right|^{\bar{n}-1}u_2(t)-\eta\left|u_2(t)\right|^{\bar n}\dot{u}_1(t)+A\dot{u}_1(t)\label{BW}
\end{array}
\end{equation}
\noindent where $u_1(t)$ represents the linear response component, and $u_2(t)$ represents the hysteretic response component. The parameters of the Bouc-Wen model are set as: $\alpha=0.1$, $\bar n=5$, $A=1$, and $\gamma=\eta=1/(2u_y^{\bar n})$, in which $u_y=0.04$ [m] is the yielding displacement. The typical hysteretic force-deformation behavior of the Bouc-Wen model is illustrated in Figure \ref{Fig2Ex3}. The initial inter-story stiffness, $\bm{k}=[k_1,k_2,k_3]$, and mass, $\bm{m}=[m_1,m_2,m_3]$, values are reported in Table \ref{tab1ex3}. The Rayleigh damping with 5 percent damping ratio for the first and second mode is adopted.
\begin{table}[h]
\centering
\caption{\textbf{Structural properties}.}
\begin{tabular}{*3c}
\toprule
{} & $k_i$[N/m] & $m_i$[kg] \\
\midrule
Story 1 & $3.0\times10^{8}$ & $1\times10^6$ \\
Story 2 & $2.8\times10^{8}$ & $1\times10^6$ \\
Story 3 & $1.5\times10^{8}$ & $1\times10^6$ \\
\bottomrule
\end{tabular}
\label{tab1ex3}
\end{table}
\noindent The horizontal forces for each story, $f_i$, $i=1,2,3$, are considered as a combination of harmonic waves with random amplitudes:
\begin{equation}
f_i(\bm{X},t)=\frac{1}{6}m_i\left(X_{1}\sin(2\pi t)+X_{2}\sin(4\pi t)+X_{3}\cos(8\pi t)+X_{4}\sin(16\pi t)\right)\,,\label{ex2Force}
\end{equation}
\noindent where $m_i$ is the mass of the $i$-th story, and $\bm{X}=[X_1,X_2,X_3,X_4]$ are independent standard Gaussian random variables. The duration of the excitation is $10$ seconds.
The model function describing the maximum absolute interstory drift is defined as
\begin{equation}
y=\mathcal{M}(\bm X)=\max_{i=1,2,3}\max_{t\in[0,10]} \left|v_i(\bm X,t)\right|\,,\label{ex2mod}
\end{equation}
\noindent where $v_1, v_2$ and $v_3$ are the first, second and third interstory drift, respectively. The CCDF of the maximum absolute drift is of practical interest, since it is an indicator of structural reliability. Therefore, with a trivial modification on the error measure (see Section \ref{Misc}), the global AL-GP method is used to estimate the CCDF. The CCDF range of interest, $[y_{\min},y_{\max}]$, is set to $[0,0.12]$. The performance of the global and conventional AL-GP approaches is reported in Table \ref{tab2ex3}\footnote{Note that in Table \ref{tab2ex3} the $\epsilon_e$ is slightly different from that in the previous examples, specifically, in Eq.\eqref{25} the denominator is replaced by $1-F_Y(y)$.}, and the estimations for the distribution mean, standard deviation, skewness and kurtosis are reported in Table \ref{tab3ex3}. The CCDF estimated via a crude Monte Carlo simulation with $10^6$ samples is used as the reference solution.
\begin{table}[H]
  \caption{\textbf{CCDF estimations averaged over 50 independent runs}.}
  \label{tab2ex3}
  \centering
  \begin{tabular}{c c c c}
    \toprule
    Methods & ${\rm E}[\epsilon_e]$ & $\sigma(\epsilon_e)$& ${\rm E}[N_{\mathcal{M}}]$\\
    \midrule
   Gaussian & 0.039 & 0.0080 & 12+124.30\\
   Dirac & 0.038 & 0.0072 & 12+115.60\\
   MoV & 0.044 & 0.0067 & 12+265.70\\
   Conventional & 0.033 & 0.0069 & 12+288.40\\
    \bottomrule
  \end{tabular}
\end{table}

\begin{figure}[H]
  \centering
    \includegraphics[scale=0.45]{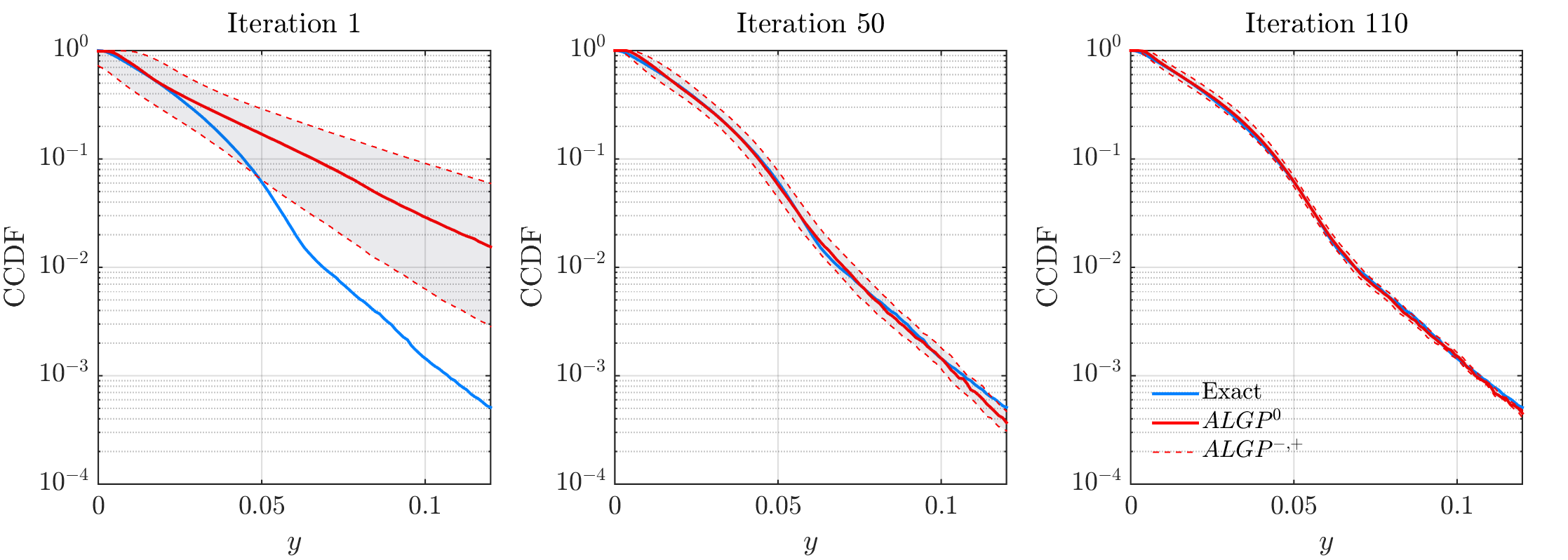}
  \caption{\textbf{Iterative process in CCDF estimation}.}\label{Fig3Ex3}
\end{figure}

\begin{figure}[H]
    \centering
    \includegraphics[scale=0.45]{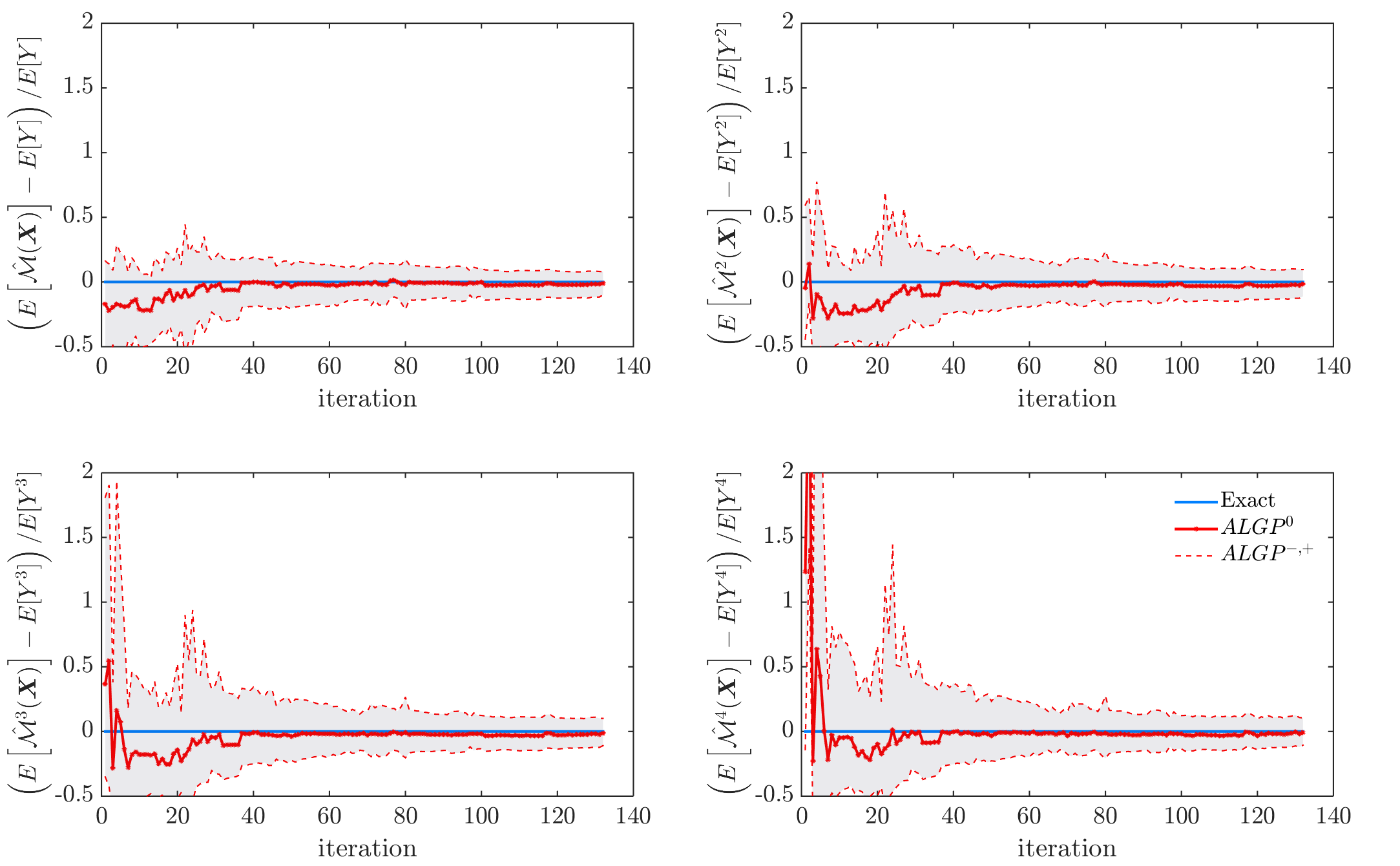}
    \caption{\textbf{Iterative process in moment estimation}.}
    \label{Fig4Ex3}
\end{figure}

\begin{figure}[H]
\centering
  \includegraphics[scale=0.45]{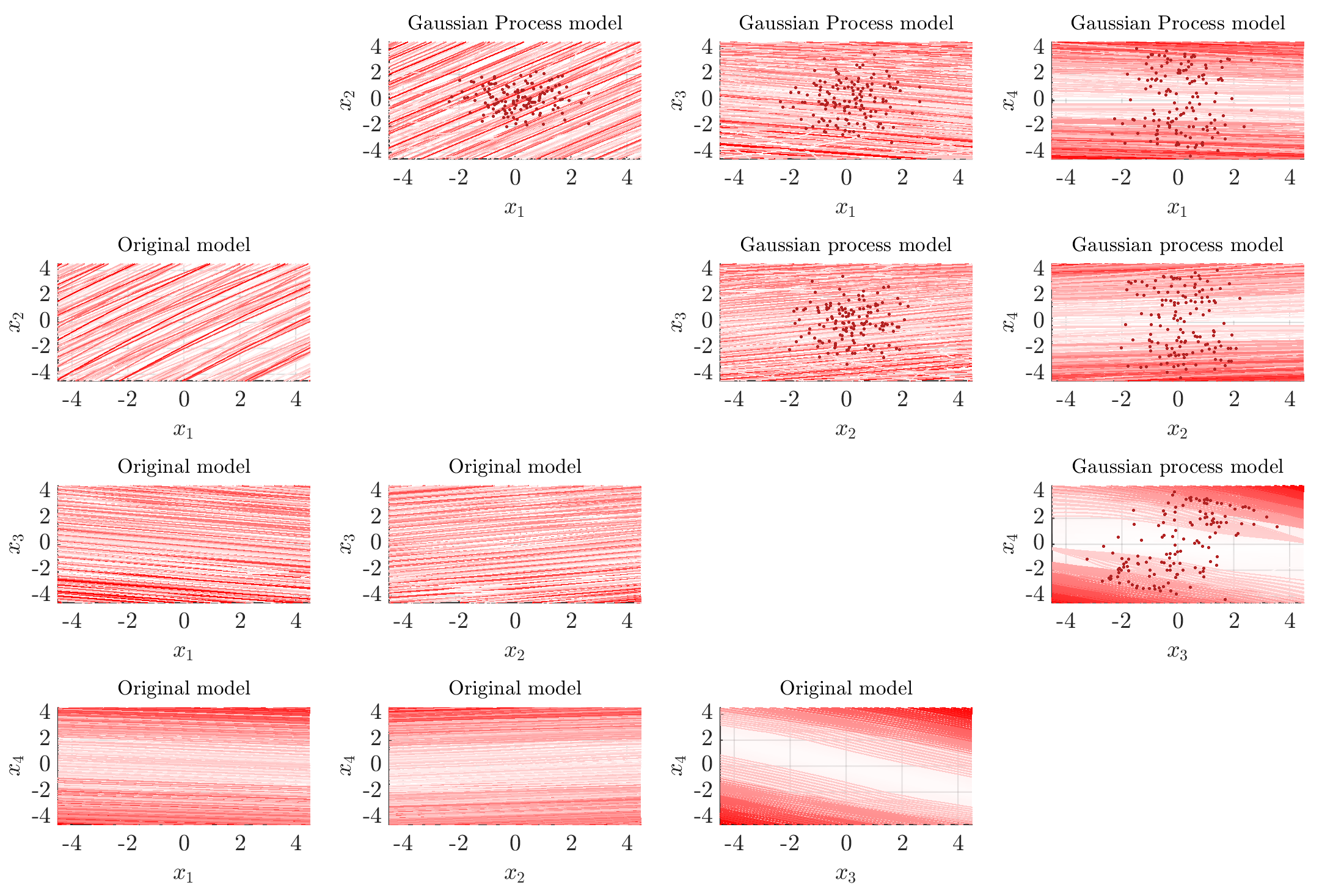}
  \caption{\textbf{Projection of the original model and the Gaussian process model in various planes}. \textit{The figure shows the projection of level sets $\left\lbrace \bm x\vert m-\mathcal{M}(\bm x)=0\right\rbrace$ and $\left\lbrace \bm x\vert m-\hat{\mathcal{M}}(\bm x)=0\right\rbrace$ in various planes. The training samples used to generate the metamodel is also shown in the figure}.}\label{Fig5Ex3}
\end{figure}

\begin{figure}[H]
\centering
  \includegraphics[scale=0.45]{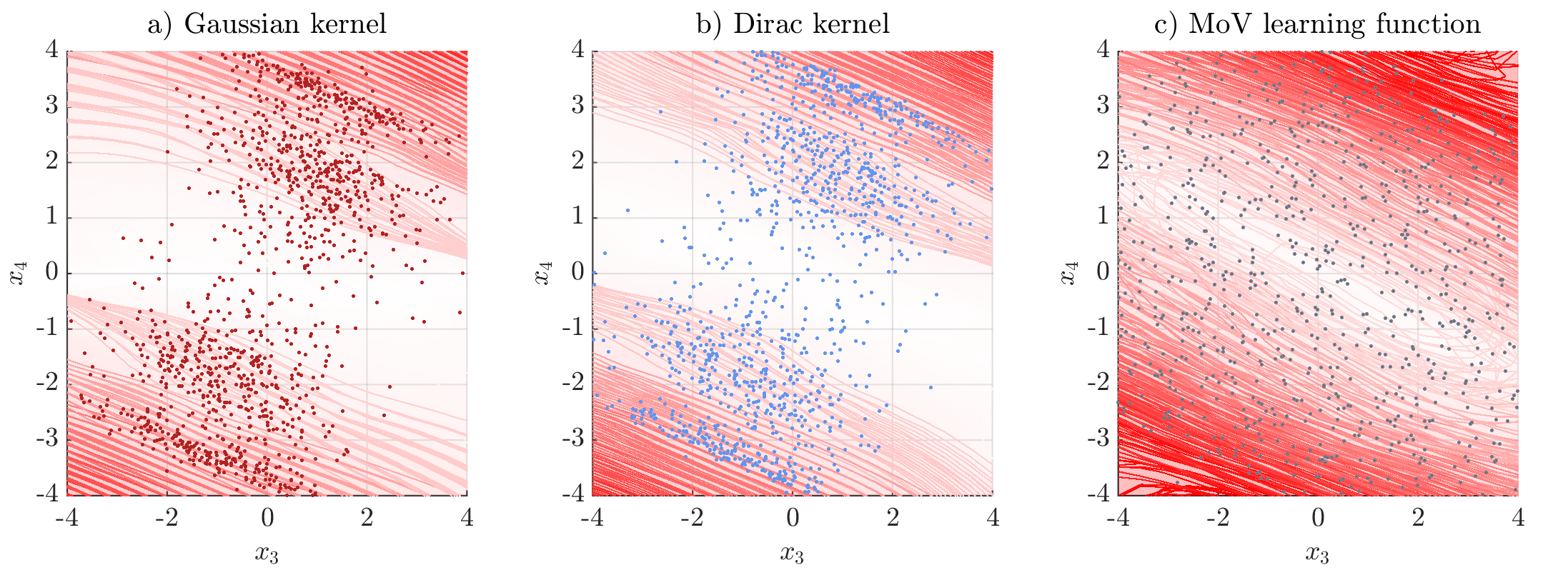}
  \caption{\textbf{Projection of the training samples in the $(x_3,x_4)$ plane}.}\label{Fig6Ex3}
\end{figure}

\begin{figure}[H]
\centering
  \includegraphics[scale=0.45]{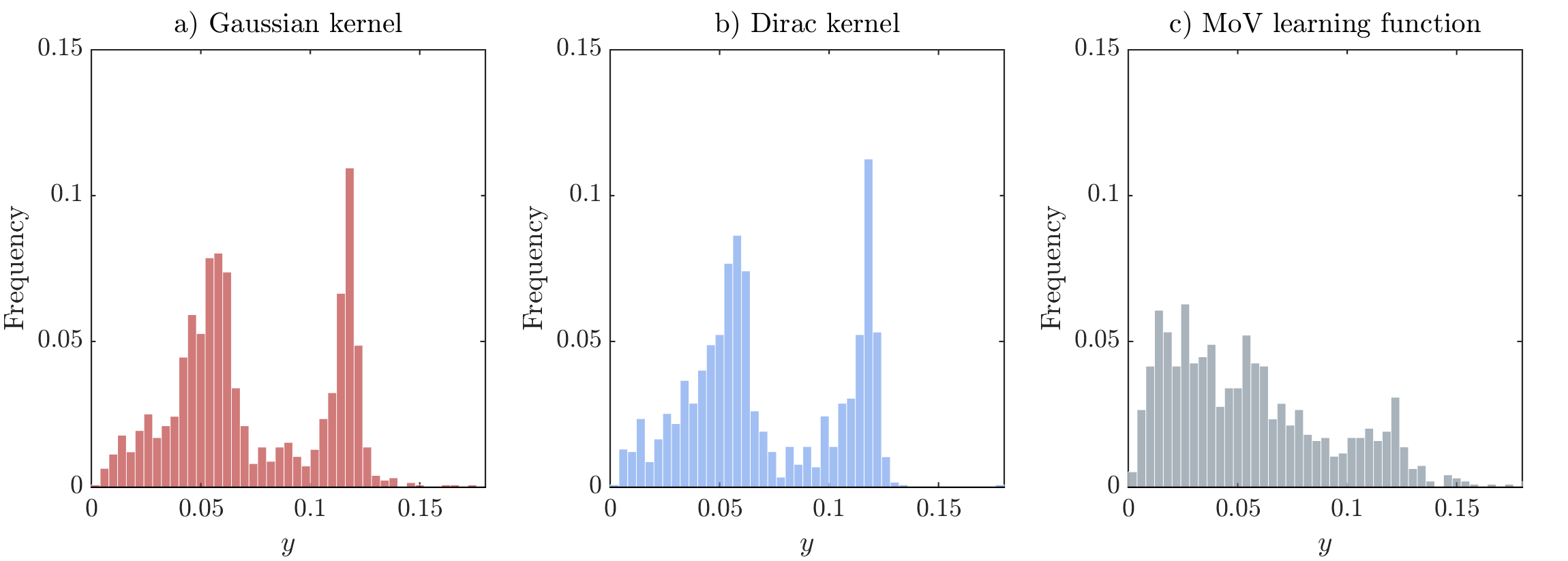}
  \caption{\textbf{Histograms on the $y$ locations of the training samples}.}\label{Fig7Ex3}
\end{figure}

\begin{table}[H]
  \caption{\textbf{Global measures of the estimated distribution function}.}
  \label{tab3ex3}
  \centering
  \begin{tabular}{c c c c c}
    \toprule
    Methods & mean/$|c.o.v.|$ & standard deviation/$|c.o.v.|$& skewness/$|c.o.v.|$ & kurtosis/$|c.o.v.|$\\
    \midrule
   Gaussian & 0.0219/0.0122 & 0.0160/0.0149 & 1.2520/0.0403 & 5.5601/0.0423\\
   Dirac &  0.0221/0.0201 &  0.0159/0.0223 & 1.2773/0.0519 & 5.6365/0.0401\\
   MoV &  0.0218/0.0150 &  0.0164/0.0168 & 1.2002/0.0446 &  5.1714/0.0446\\
   Conventional &  0.0222/0.0055 &  0.0159/0.0069 & 1.2603/0.0334 &  5.5753/0.0394\\
   Monte Carlo & 0.0221  & 0.0160 & 1.2256   & 5.4799\\
    \bottomrule
  \end{tabular}
\end{table}
\noindent
Figure \ref{Fig3Ex3} and Figure \ref{Fig4Ex3} show typical iterative processes for the CCDF and moment estimations from the global AL-GP method with Gaussian kernel. Figure \ref{Fig5Ex3} shows projection of the original model and the Gaussian process model in various planes. Figure \ref{Fig6Ex3} shows projection of the training samples in the $(x_3,x_4)$ plane. Figure \ref{Fig7Ex3} shows histograms of the $y$ locations of the training samples. 
\noindent
It is seen from Figure \ref{Fig7Ex3} that the training samples are concentrated around the tail and the transition region where the stiffness degrades significantly.

\section{Practical issues, limitations, and future directions}\label{Misc}


\noindent One practical issue that deserves attention is the stopping criterion Eq.\eqref{StopC}. In particular, it should be compatible with the capability of the specified Monte Carlo simulation technique. For example, if a crude Monte Carlo simulation with $10^6$ samples is used in Step 3 of Algorithm \ref{alg:01}, and we intend to estimate a CDF/CCDF value as low as $10^{-4}$, then the coefficient of variation of using $10^6$ samples is circa $10^{-1}
$. Consequently, the $\bar \epsilon$ in Eq.\eqref{epsilon} should be specified larger than 0.1, otherwise the tolerance is unnecessarily tight (which requires unnecessarily more model function evaluations).

A practical, albeit trivial, issue is that for some applications only the CDF \emph{or} the CCDF (e.g., Example 3) needs to be accurately estimated. For this case, one can simply replace the $\min[\hat{F}_Y^0(y),1-\hat{F}_Y^0(y)]$ term in error measures $W^*$ and $W_L^*$ with $\hat{F}_Y^0(y)$ or $1-\hat{F}_Y^0(y)$, and the rest part of the algorithm remains intact. The same holds for outputs with symmetric probability distribution (which is known a-priori). Observe that in Example 2, we compute both tails because we assumed no a-priori knowledge on the shape of $f_Y(y)$.

We mention also an important issue for future studies: the existence of bifurcation/singularity in the model function. Suppose the model function has singularity at $y_s=\mathcal{M}(\bm x)$, i.e. the behavior of $M(\cdot)$ for $y<y_s$ is fundamentally different from the behavior for $y>y_s$, and the transition is abrupt. If the three-fold metamodel approach (see Eq.\eqref{9}) is used to estimate the probability around $y_s$, the $\hat{\mathcal{M}}^+$ and $\hat{\mathcal{M}}^-$ model may span different sides of $y_s$, and consequently an error measure using the discrepancy between $\hat{\mathcal{M}}^+$ and $\hat{\mathcal{M}}^-$ could be extremely large regardless of numerous training samples may already be applied to the singularity region. In the context of distribution function estimation, if $y_s\in[y_{\min},y_{\max}]$, the algorithm would keep applying training samples around $y_s$ yet it could hardly converge. A possible solution to this issue is to use transformation techniques such that in the projected feature space the singularity may be smoothed out \cite{ManifoldGP}\cite{WarpedGP}.

 It is also important to note that the proposed active learning approach for distribution function estimation is not restricted to Gaussian process model. In fact, the approach can be used with any metamodels as long as the model error/uncertainty can be evaluated. This idea instantly generates interesting research topics for future studies, e.g. the use of polynomial chaos expansion (with bootstrap technique) in distribution function estimation \cite{PCEboot}.
 
 Finally, the distribution function estimation for high dimensional model functions can be investigated using classical and/or nonlinear dimensionality reduction techniques, and manifold embedding. 

\section{Conclusions}\label{Conclusion}

\noindent This study proposes a global active learning-based Gaussian process metamodelling strategy for estimating the probability distribution function in forward  uncertainty quantification analysis. The strategy is mesh free in the sense that a-priory discretization (mesh) of the distribution function is not required. A novel  error measure is developed such that it satisfies a symmetric condition between cumulative and complementary cumulative distribution functions. As a result of this symmetry, the proposed method is able to simultaneously provide accurate CDF and CCDF estimation in their median-low probability regions. Moreover, a two-step learning function is proposed such that it makes full use of the available information and it is compatible with the error measure. The proposed metamodelling strategy has been tested for three benchmark examples showing both high accuracy and efficiency.


\section*{Acknowledgement}
\noindent
Dr. Ziqi Wang was supported by the National Science and Technology Major Project of the Ministry of Science and Technology of China (Grant No. 2016YFB0200605), National Natural Science Foundation of China (Grant No.1808149), and the Natural Science Foundation of Guangdong Province (Grant No.2018A030310067). Dr. Marco Broccardo was supported by the Swiss Seismological Service and by the Chair of Structural Dynamics and Earthquake Engineering (both) at the Swiss Federal Institute of Technology (ETH). This support is gratefully acknowledged. Any opinions, findings, and conclusions expressed in this paper are those of the authors, and do not necessarily reflect the views of the sponsors.





\bibliography{ALCDF}







\end{document}